\documentclass[final,1p,times]{elsarticle}
\usepackage{graphics}
\usepackage{subfigure}
\usepackage{epstopdf}
\usepackage[figuresright]{rotating}
\usepackage{multicol}
\usepackage{color}
\usepackage{amsmath, amssymb}
\usepackage{amssymb}
\usepackage{amsthm}
\usepackage[ruled, linesnumbered]{algorithm2e}
\usepackage{algorithmic}
\usepackage{setspace}
\usepackage{multirow}
\usepackage{lineno}
\usepackage{array}
\usepackage{booktabs}
\usepackage{amsmath}
\newtheorem{theorem}{Theorem}

\newtheorem{definition}{Definition}
\usepackage{algorithm2e,setspace}
\graphicspath{{kbsGraphs/}}
\usepackage{pdfpages}

\newtheoremstyle{noparens}%
  {}{}%
  {\itshape}{}%
  {\bfseries}{.}%
  { }%
  {\thmname{#1}\thmnumber{ #2}\mdseries\thmnote{ #3}}
\theoremstyle{noparens}
\newtheorem{theoremNoParens}[theorem]{Theorem}

\journal{under review}
\begin{document}
\begin{sloppypar}
\begin{frontmatter}
\title{Learning Causal Representations for Robust Domain Adaptation}
\author[1,2]{Shuai Yang}
\ead{yangs@mail.hfut.edu.cn}
\author[1,2]{Kui Yu}
\ead{yukui@hfut.edu.cn}
\author[3]{Fuyuan Cao}
\ead{cfy@sxu.edu.cn}
\author[4]{Lin Liu}
\ead{Lin.Liu@unisa.edu.au}
\author[1,2]{Hao Wang}
\ead{ jsjxwangh@hfut.edu.cn}
\author[4]{Jiuyong Li}
\ead{Jiuyong.Li@unisa.edu.au}

\address[1]{ Key Laboratory of Knowledge Engineering with Big Data (Hefei University of Technology), Ministry of Education}
\address[2]{ School of Computer Science and Information Engineering, Hefei University of Technology, Hefei 230009, China}
\address[3]{ School of Computer and Information Technology, Shanxi University, Taiyuan 030006, China}
\address[4]{ UniSA STEM,  University of South Australia, Adelaide, SA 5095, Australia}

\begin{abstract}
Domain adaptation solves the learning problem in a target domain by leveraging the knowledge in a relevant source domain. While remarkable advances have been made, almost all existing domain adaptation methods heavily require large amounts of unlabeled target domain data for learning domain invariant representations to achieve good generalizability on the target domain. In fact, in many real-world applications, target domain data may not always be available. In this paper, we study the cases where at the training phase the target domain data is unavailable and only well-labeled source domain data is available, called robust domain adaptation. To tackle this problem, under the assumption that causal relationships between features and the class variable are robust across domains, we propose a novel Causal AutoEncoder (CAE), which integrates deep autoencoder and causal structure learning into a unified model to learn causal representations only using data from a single source domain. Specifically, a deep autoencoder model is adopted to learn low-dimensional representations, and a causal structure learning model is designed to separate the low-dimensional representations into two groups: causal representations and task-irrelevant representations. Using three real-world datasets the extensive experiments have validated the effectiveness of CAE compared to eleven state-of-the-art methods.
\end{abstract}
\end{frontmatter}
\section{Introduction}
\label{Introduction}
\par In traditional machine learning, a fundamental assumption is that training data (source domain) and testing data (target domain) are drawn from the same distribution. However, in many practical scenarios, such as image classification, news recommendation and sentiment analysis, this assumption is often violated ~\cite{zhuangsurvey,RenXY20,ChenSLYW19}.

\par To tackle this problem,  a variety of domain adaptation methods have been proposed \cite{WeiK19MIMT,LC20,DengLZ19,LXSDH19,Kang0YH19}. Assuming a source domain and a relevant target domain which have the same feature space but different probability distributions, domain adaptation aims to learn a set of domain invariant features  that can generalize well to the target domain. Existing domain adaptation methods can be roughly grouped into two categories based on the approach taken: correlation-based  methods, such as CDAN~\cite{LongC0J18} and ALDA~\cite{LC20}, and causality-based  methods, such as  Infer~\cite{kunZhang2020} and Prop~\cite{prop}.


\par  Correlation-based  methods learn invariant representations based on the correlations between features and the class variable. In general, correlations may not capture the causal relationships between features and the class variable, but only their co-occurrences, and thus correlation-based invariant representations may not guarantee good generalizability on the target domain. For instance, in a picture of a dog, those representations of the background features such as \emph{grass} and \emph{blue sky} may have a strong correlation with the label (dog), but they may not be transferable across different target domains,  e.g. pictures taken at different locations or weather conditions, as they are not the causal features of a dog image.


\par To mitigate this problem, learning  invariant causal representations between source and target domains  from data has attracted much attention recently \cite{kunZhang2020,prop}. Causal relationships reflect the underlying data generating mechanism and thus  are persistent across different settings or environments. For instance, considering the picture of a dog again, features such as \emph{eyes}, \emph{ears} and \emph{face shape} are the fundamental elements for recognizing a dog and they would not change no matter where and when the picture was taken.

\par Although  domain adaptation has been studied quite intensively in recent years, most existing domain adaptation methods  rely on the unlabeled target domain data for learning domain invariant representations \cite{zhuangsurvey}.  Indeed, the target domain data provide an additional source of information  for learning invariant representations with better generalizability. However, in practice, often only a very limited amount of target domain data is available, either because the target domain is evolving (e.g. an online movie review system for which new movies are produced all the time), or it is impossible to collect all samples in the target domain (e.g. a computer vision system for which it is impossible to collect samples for all different contexts). Therefore classifiers trained on the  representations learnt by the existing methods may generalize well to the very small portion of the known target domain data, but their performance may deteriorate soon after being deployed in real life.

\par To alleviate the  problem mentioned above, zero-shot domain adaptation (ZSDA) has been studied \cite{WangJ19,ZSDA,ALZSDA}, which deals with the situation where the task-relevant target domain data is unavailable but the task-irrelevant source and target domain data is available during the training phase. For instance, ZSDA methods learn a prediction model for the color digit images (task-relevant target domain), given the grayscale digit images (task-relevant source domain), the grayscale letter images (task-irrelevant source domain), and the color letter images (task-irrelevant target domain). Existing methods  \cite{ZSDA,WangJ19,ALZSDA} depend on the task-irrelevant target domain data for learning domain invariant representations to bridge the gap between source and target domains. However, in  real-world scenarios, on the one hand, collecting sufficient task-irrelevant target domain data is  expensive and time-consuming. On the other hand, when the  difference between the collected  task-irrelevant target domain data and  task-relevant target domain data is substantially large, matching the distributions between the task-irrelevant dual-domain pairs may not well alleviate the distribution discrepancy between task-relevant source domain data and task-relevant target domain data.



\par Then a question naturally arises: when the task-relevant target domain data is inadequate or unavailable and the task-irrelevant target domain data is unavailable, can we learn domain invariant feature representations with good generalizability only using data of the  source domain?

\par In this paper, we formulate this question as the robust domain adaptation problem for cases where the target domain data (both task-relevant and task-irrelevant target domain data) is unavailable and only a well-labeled source domain is available in the training phase. The key challenge for robust domain adaptation is  how to reduce the distribution discrepancy between the source and unknown target domains.
 \par Recent studies reveal that causal features enable more reliable predictions in non-static environment where the distributions of training and  testing data may be different but related \cite{CausalityML,yu2020causality,RCIT}. Using causal features of the class variable to built a prediction model  is a reasonable way to address the problem of robust domain adaptation, since even if the source and target domains  are not obtained from the same  data distribution, the causal relationships between features and the class variable are robust across domains. For instance, we want to build a  predictive model  for   recognizing dogs using the images with dogs on  grass. Based on the images  with dogs on  grass, a prediction model built using non-causal features such as \emph{grass} may not achieve good performance on the image with a dog in snow. In contrast, if the causal features (such as \emph{eyes}, \emph{ears} and \emph{face shape}) are selected as the predictive features, a predictive model built on the images with dogs on  grass will be robust to correctly classify an image with a dog in snow.  Recently, a number of Markov Blanket (MB) discovery algorithms have been proposed for causal feature selection \cite{BAMB,yu2020causality}. The MB of the class variable consists of the parents (direct causes), children (direct effects), and spouses (other parents of the class variable's children). For more details of MB, we refer readers to Appendix of this paper and related literature, e.g. \cite{yu2020causality,BAMB,S2TMB}.  Due to the robustness of causal features \cite{CausalityML,RCIT}, we can learn MB of the class variable for robust domain adaptation. However, high-dimensional data often deteriorate the performance of MB learning, due to the fact that when dimension increases, MB learning becomes unreliable given a limited number of samples. Furthermore,  noise contained in  data will make existing MB learning algorithms achieve inaccurate results.
%

 \par Motivated by the aforementioned issues, in this paper, under the assumption that causal relationships between features and the class variable are robust across domains, we propose a novel Causal AutoEncoder (CAE) for learning robust causal representations only using data from a single source domain. Specifically, CAE jointly optimizes deep autoencoder and causal structure learning models to learn causal  representations  from the source domain data. The deep autoencoder model is adopted to map the input data into a low-dimensional feature space to reduce the influence of  noise contained in  data, and the causal structure learning model is used to separate the low-dimensional representations learnt with the deep autoencoder into two groups: causal representations (MB feature representations) and task-irrelevant representations (non-MB feature representations). The main contributions of the paper are summarized as follows.


\begin{itemize}
\item  We investigate the robust domain adaptation problem in which the task-relevant and task-irrelevant target domain data is unknown during the model training phase.
\item  For robust domain adaptation, we design a causal autoencoder, which integrates autoencoder and causal structure learning   into a unified  model to  learn  causal representations.
\item  We have conducted extensive experiments using three public domain adaptation datasets, and have compared CAE with eleven  state-of-the-art algorithms, to demonstrate the effectiveness and  superiority of CAE.
\end{itemize}

\section{Related Work}
\label{relatedwork}
\par Our work is  in the area of domain adaptation and is also related to domain generalization  and stable learning.  This section reviews the related work in the three areas and briefly introduces causal feature selection.

\par \textbf{Domain adaptation.} Existing domain adaptation methods  fall into two  broad categories based on the approach taken: correlation-based  and causality-based  methods.
\par Correlation-based  domain adaptation methods  have received increasing attentions from researchers  because of their impressive performance \cite{LLXDHT20,SSRLDA}.   Generally, these  methods can be autoencoder-based or adversarial-learning based. One typical autoencoder-based method is DTFC, which  disturbs different features with  different corruption probabilities to extract complex  feature representations \cite{WeiFeature}. Nevertheless, DTFC only depends on the single autoencoder model, which presents a problem when  extracting multiple characteristics of data. To alleviate this problem, SERA \cite{SERA} and SEAE \cite{SEAE} tack two types of autoencoders in series to capture different characteristics of data in both domains. One typical adversarial-learning method is CDAN \cite{LongC0J18}, which trains deep networks in a conditional domain-adversarial paradigm. To improve the transferability of deep neural networks, TransNorm  \cite{WangJLWJ19} is presented. ALDA  \cite{LC20} integrates domain-adversarial learning and self-training into a unified framework to learn discriminative feature representations for accomplishing cross-domain tasks.
 \par ZSDA \cite{ZSDA} and CoCoGAN \cite{WangJ19}  focus on tackling the zero-shot domain adaptation  problem, where  the target-domain data for the task of interest is unavailable but the   task-irrelevant dual-domain pairs data is available.  Both methods rely on  the  information from task-irrelevant dual-domain pairs to complete cross-domain tasks. Our method  differs from ZSDA  and CoCoGAN  because we only use the data from a single source domain and  need not leverage the  information from task-irrelevant dual-domain pairs.


%

\par  Recently, several  causality-based  methods have been designed for domain adaptation, such as Infer~\cite{kunZhang2020} and Prop~\cite{prop}.  Infer ~\cite{kunZhang2020} uses a graphical model as a compact way to encode the change property of the joint distribution and then treats domain adaptation as a problem of Bayesian inference on the graphical models.  Prop ~\cite{prop} captures  the underlying data  generating  mechanism  behind the data distributions to accomplish cross-domain tasks.

\par We argue that existing domain adaptation methods leverage the target domain information or the information from task-irrelevant dual-domain pairs, but our method only  uses the information from a single source domain.

\par \textbf{Domain generalization.}  Domain generalization focuses on leveraging the knowledge in multiple related domains to promote the learning task in unseen domains \cite{GhifaryKZB15}. A variety of  domain generalization approaches have been designed, such as MTAE \cite{GhifaryKZB15}, MMD-AAE \cite{LiPWK18} and MASF \cite{DouCKG19}. MTAE \cite{GhifaryKZB15} uses multi-task autoencoder to  learn  invariant features that are robust cross domains.  MMD-AAE \cite{LiPWK18}  learns latent feature representation by jointly optimizing a multi-domain autoencoder regularized by a classifier, the maximum mean discrepancy  distance and  an discriminator in an adversarial training manner. MASF \cite{DouCKG19} preserves  inter-class relationships by aligning a derived soft confusion matrix, and  encourages domain-independence while separating sample features and class-specific cohesion by using a metric-learning component. Recently, some causality-based methods have been proposed for domain generalization ~\cite{MagliacaneOCBVM18,RojasCarullaST18}.  We argue that existing domain generalization methods rely on multiple source domains to explore the invariance properties, whereas our method only uses the data of a single source domain.

\par \textbf{Stable learning}. Stable learning deals with the challenging situation where only a single training data  is available and the testing data is unavailable during the training phase \cite{DGBR,Shen0ZK20}. Stable learning focuses on tackling the problem of  selection bias and model misspecification bias. Recently, several  approaches  have been designed for tackling  sample selection bias problem, such as CRLR \cite{CRLR} and DGBR \cite{DGBR}. CRLR  and DGBR balance the distribution of  treated and control group  by  global sample weighting to isolate the clear effect of each predictor from the confounding variables. Despite their better performance,  learning the weight for each sample is difficult in big data scenarios with huge training samples. To alleviate this issue, CVS \cite{RCIT} separates the causal variables by leveraging a seed variable. To deal with  the problem of sample selection bias and model misspecification bias simultaneously, Shen et al. \cite{Shen0ZK20}  reduce collinearity among input variables by sample  reweighting, and Kuang et al. \cite{KuangX0A020} force all features to be  as independent as possible by jointly optimizing  a weighted regression model and a variable decorrelation regularizer. These methods  either reweight each sample or select causal features for stable learning, whereas our method  learns causal representations for domain adaptation by  jointly  optimizing deep autoencoder and causal structure learning models.

\par \textbf{Causal feature selection.} Feature selection plays an essential role in high-dimensional learning tasks and it has been widely used in various  practical  scenarios.   Traditional correlation-based feature selection methods  exploit  statistical associations or dependences  between features and the class variable to make predictions, such as FCBF \cite{YuL04} and mRMR \cite{PengLD05}. Correlation-based feature selection methods may capture  spurious correlation, leading to unstable prediction across different environments.  Recently, causal feature selection  has attracted much attention from researchers \cite{yu2020causality}, as it not only can improve the explanatory capability of predictive models  but also can improve the robustness of predictive models. Learning Markov blanket (MB) for causal feature selection  has been extensively studied in recent years, since the MB of the class variable is the optimal feature subset for feature selection \cite{TsamardinosA03}.   Existing MB discovery methods can be roughly grouped into constraint-based methods and score-based methods. The former learns the MB of the class variable using conditional independence tests, such as HITON-MB \cite{AliferisTS03} and BAMB \cite{BAMB},  while the latter  employs Bayesian score functions to discover  MB instead of using conditional independence tests, such as  SLL  \cite{NiinimakiP12} and S$^2$TMB  \cite{S2TMB}. Constraint-based MB discovery algorithms cannot distinguish direct causes of a target variable from its direct effects.  Score-based algorithms can distinguish parents from children of a given variable, but they may suffer  high computational complexity. For more details of MB,  please refer to the Appendix.



\section{The Proposed Causal AutoEncoder}
In this section, we first  formulate the robust domain adaptation problem to be tackled in this paper, and then introduce the proposed CAE algorithm in detail.

\subsection{Problem Formulation}

\textbf{Robust Domain Adaptation}. Given a labeled source domain $\mathcal{D}_{s} = [\textbf{X}_{s},\textbf{y}_{s}]$, where $ \textbf{X}_s \in  \mathbb{R}^{n\times d}$ and  $\textbf{y}_s = (y_1, ..., y_n)^T \in  \mathbb{R}^{n\times 1}$  represent the source domain data of the features and class label $Y$, respectively. $n$  denotes the number of samples and $d$  denotes the dimension of feature space,  $y_{i}$ is the label of the $i^{th}$ sample. Robust domain adaptation aims to leverage the  knowledge in only one single source domain to learn a robust predication model for  predicting the label of  target domain data $\mathcal{D}_{t} = \textbf{X}_{t}$.


\begin{figure*}
\centering
\subfigure{
    \includegraphics[ width=14cm]{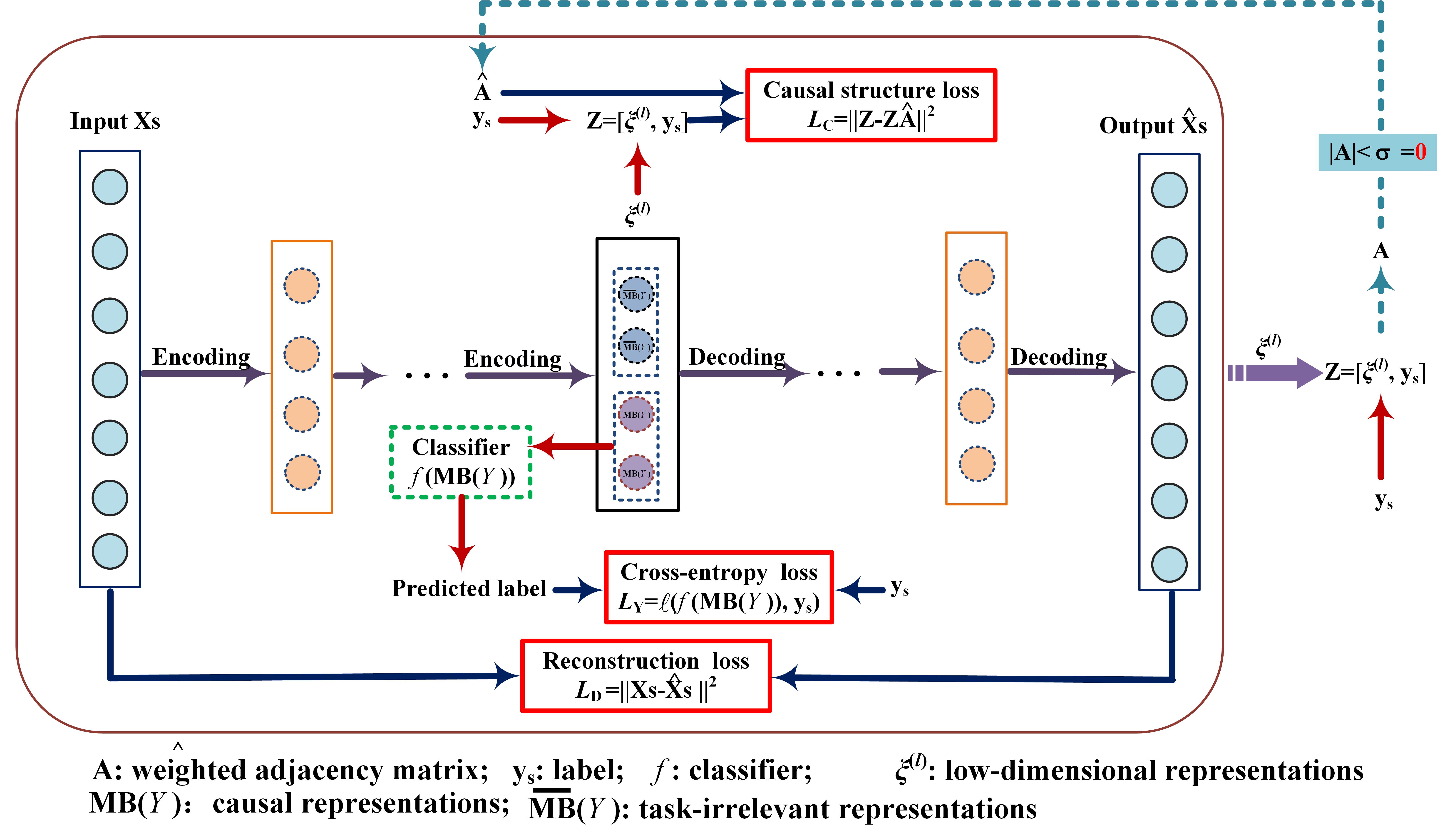}
}
\caption{The framework of  CAE.}
 \label{FrameworkCAE}
\end{figure*}


\subsection{Overview of CAE}

\par We propose the Causal AutoEncoder (CAE) algorithm to learn  causal representations for robust domain adaptation. The  framework of CAE is illustrated in Fig.\ref{FrameworkCAE}.  CAE has an input layer, an output layer and multiple hidden layers. Different from the traditional autoencoder, CAE incorporates the causal structure learning loss $\mathcal{L}_{C}$ and cross-entropy loss $\mathcal{L}_{Y}$ to learn robust causal low-dimensional  representations with high generalizability for unknown target domain. The objective function to be minimized in CAE can be summarized as follows.

\begin{equation}
\footnotesize
\begin{aligned}
\mathcal{L} = \mathcal{L}_{D} + \lambda_1 \mathcal{L}_{C} + \lambda_2\mathcal{L}_{Y} + \lambda_3 \mathcal{L}_{R}.
\label{LossCAE}
\end{aligned}
\end{equation}
where  $\lambda_1$, $\lambda_2$  and $\lambda_3$  are the balancing parameters.
 \par CAE learns low-dimensional  representations by minimizing the construction error $\mathcal{L}_{D}$ between the input data $\textbf{X}_s$ and output data $\hat{\textbf{X}}_s$. To learn causal  representations, CAE uses causal structure learning loss $\mathcal{L}_{C}$ to   force the low-dimensional  representations  into two different groups, i.e.  causal  representations $\textbf{MB}(Y)$ and task-irrelevant  representations $\overline{\textbf{MB}}$$(Y)$, as shown in Fig.\ref{FrameworkCAE}. The  cross-entropy loss  $\mathcal{L}_{Y}$  is used to optimize  causal  feature representations by incorporating the label information of source domain data. $\mathcal{L}_{R}$ is a regularization term on model parameters.  In the following, we will give the details of CAE.

\subsection{Reconstruction loss $\mathcal{L}_{D}$ and regularization loss $\mathcal{L}_{R}$}
\par  Recent studies  reveal that even if the source and target domains are not obtained from the same data distribution,  the causal relationships between features and the class variable  are robust across relevant domains under different environments or settings \cite{RCIT,DGBR,CRLR}. Based on this,  learning  causal features of the class variable is a  reasonable way to address the problem of robust domain adaptation. However, high-dimensional data  often deteriorate the performance of  causal feature learning. On the one hand, causal feature learning becomes unreliable on  high-dimensional data given a limited number of samples. On the other hand,  noise contained in  data can lead to inaccurate results of  existing causal feature learning algorithms. To alleviate  these problems,  CAE adopts a deep autoencoder model to map the input data into a low-dimensional feature space with high-level representations. CAE consists of two stages: encoding and decoding. Given  input data $\textbf{X}_s$, CAE first encodes the input data to  learn the low-dimensional feature  representations through multiple nonlinear encoding processes, then decodes the low-dimensional representations to obtain the estimated output data $\hat{\textbf{X}}_s$.  The process can be formalized as follows.
\begin{equation}
\footnotesize
\begin{aligned}
&{\rm Encoding:}\; \boldsymbol{\xi}^{(j)} = \sigma(\boldsymbol{\xi}^{(j-1)} \textbf{W}_{1}^{(j)} + \textbf{b}_{1}^{(j)}), j=1,2,\cdots,l.\\
&{\rm Decoding:}\; \boldsymbol{\psi}^{(j)} = \sigma(\boldsymbol{\psi}^{(j-1)} \textbf{W}_{2}^{(j)} + \textbf{b}_{2}^{(j)}), j=1,2,\cdots,l.
\label{DAEEnDE}
\end{aligned}
\end{equation}
where $l$ is the number of hidden layers in the autoencoder and $\sigma$ is a nonlinear activation function (e.g., sigmoid function). Note that $\boldsymbol{\xi}^{(1)}$ = $\textbf{X}_s$ and $\boldsymbol{\psi}^{(1)}$ = $\boldsymbol{\xi}^{(l)}$, and $\boldsymbol{\xi}^{(l)} \in \mathbb{R}^{n\times k}$ ($k<d$ ) is the low-dimensional representations for $\textbf{X}_s$. $\textbf{W}_{1}^{(j)}$ and $\textbf{b}_{1}^{(j)}$ are the weight matrix and the bias vector on the $j^{th}$ encoding layer, respectively. For the $j^{th}$ decoding layer, $\textbf{W}_{2}^{(j)}$ and $\textbf{b}_{2}^{(j)}$ are its weight matrix and  bias vector, respectively.  The reconstruction error $\mathcal{L}_{D}$ (see Eq.(\ref{LossCAE})) between the input data  $\textbf{X}_s$ and  output data $\hat{\textbf{X}}_s$ = $\boldsymbol{\psi}^{(l)}$ in Eq.(\ref{LossCAE}) is defined as follows.
\begin{equation}
\footnotesize
\begin{aligned}
\mathcal{L}_{D} = \frac{1}{2n}\|\textbf{X}_s-\hat{\textbf{X}}_s\|^2.
\label{DAE}
\end{aligned}
\end{equation}

\par The last term in  Eq.(\ref{LossCAE}), i.e. $\mathcal{L}_{R}$, is a regularization term on model parameters, which is calculated as follows.
\begin{equation}
\footnotesize
\begin{aligned}
\mathcal{L}_{R} = \sum_{j=1}^{l}(||\textbf{W}_{1}^{(j)}||^2 + ||\textbf{b}_{1}^{(j)}||^2+||\textbf{W}_{2}^{(j)}||^2 + ||\textbf{b}_{2}^{(j)}||^2).
\label{CAER}
\end{aligned}
\end{equation}

\subsection{Causal structure learning loss  $\mathcal{L}_{C}$}
\par  In order to reconstruct the input data  $\textbf{X}_s$, the intrinsic information of input data needs to be captured by the low-dimensional feature representations $\boldsymbol{\xi}^{(l)}$. Therefore,  $\boldsymbol{\xi}^{(l)}$ may contain some task-irrelevant information that cannot be used for cross-domain tasks, such as the background information in an image. Thus we hope to  separate the low-dimensional representations into two groups:  causal  representations and task-irrelevant  representations. Studies  \cite{yu2020causality,neufeld1993pearl} reveal that the Markov blanket (MB) of the class variable is the minimal feature subset with maximum predictivity for classification.  Under the faithfulness assumption \cite{CPS},  all other features that are not in  the MB set of the class variable are conditionally independent of  the class variable conditioning on  the MB  set of the class variable. This means that the information of the non-MB feature representations are blocked given the MB feature representations.  Motivated by this, in this paper, we treat  the MB feature representations of the class variable as causal representations and other feature (non-MB features) representations  as  task-irrelevant  representations, since the information of task-irrelevant  representations are blocked by the  MB feature representations. Thus,  we need to identify MB feature representations of the class variable.
\par  To achieve this goal, based on the learnt low-dimensional representations $\boldsymbol{\xi}^{(l)}$ and label information $\textbf{y}_s$ of source domain, we  aim to learn a directed acyclic graph (DAG) $\mathbb{G}$ over $\textbf{Z}$, where $\textbf{Z}=[\boldsymbol{\xi}^{(l)},\textbf{y}_s] \in \mathbb{R}^{n\times(k+1)}$. In  $\mathbb{G}$, $Z_i \in \textbf{Z}$ is a parent (i.e. a direct cause) of $Z_j$ and $Z_j$ is a child (i.e. a direct effect) of $Z_i$ if there is a directed edge from variable $Z_i$ to variable $Z_j$. Once $\mathbb{G}$ is obtained, we can identify the MB features of the class variable.


\par In the following, we give the details of learning a DAG $\mathbb{G}$ over $\textbf{Z}$. In this paper,  we assume the same data generating procedure as in \cite{ZhengARX18}, i.e. the values of a variable is generated based on its direct causes, that is,
\begin{equation}
\footnotesize
Z_i = g(Z_{pa(i)})+N_{i}, i = 1,2,\cdots,k+1.
\label{SEM}
\end{equation}
where $Z_{pa(i)}$ represents the set of parent  variables of variable $Z_i$, $g$ is a mapping function and $N_{i}$ is  an additive noise.

\par Inspired by  Zheng et al. \cite{ZhengARX18}, we encode the DAG $\mathbb{G}$ with the set of variables \textbf{Z} (i.e. the $k$-dimensional high-level feature representations and the class variable $\textbf{y}_s$) using an adjacency matrix $\textbf{A}$ = $[\textbf{a}_{1}|\cdots|\textbf{a}_{k+1}]\in\mathbb{R}^{(k+1)\times (k+1)}$, where each column $\textbf{a}_i$ denotes the coefficients in the linear  structural equation model  of the form $Z_i$ = $\textbf{a}_{i}^{T}\textbf{Z}+N_{i}$ , or \textbf{Z} = \textbf{ZA} + \textbf{\emph{N}}, where \textbf{ ZA} is the estimated value for \textbf{Z} based on \textbf{Z}'s parents and we want the estimation as accurate as possible.  To ensure the acyclicity of $\textbf{A}$,  a  constraint on $\textbf{A}$ is adopted,  as indicated in  Theorem \ref{theorem_acy} below.
\begin{theorem}\label{theorem_acy}
Let \textbf{A}   $\in\mathbb{R}^{(k+1)\times (k+1)}$ be the (possibly negatively) weighted adjacency matrix of a directed graph, the graph is acyclic if and only if
\begin{equation*}
\footnotesize
tr\big[\big(\textbf{I}+\frac{\textbf{A} \odot\textbf{A}}{1+k}\big)^{(k+1)}\big]-(k+1)=0.
\label{acyclicity2}
\end{equation*}
\end{theorem}
where $\odot$ is the Hadamard product, $\textbf{I}$ is the  the unit matrix.
\begin{proof}
Let \textbf{B} = \textbf{A} $\odot$ \textbf{A}. Note that \textbf{B} is nonnegative. The binomial expansion reads
\begin{equation*}
\footnotesize
\big(\textbf{I}+\frac{\textbf{B}}{1+k}\big)^{(k+1)} = \textbf{I} + \sum_{j=1}^{k+1} \binom{k+1}{j} (\frac{1}{1+k})^{j}\textbf{B}^{j}.
\label{acyclicity}
\end{equation*}
It is known that there is a cycle of length $j$ if and only if $tr(\textbf{B}^{j})>0$ when \textbf{B} $\geq$ 0.  Because if there is a cycle then
there is a cycle of length at most $k+1$, we conclude that there is no cycle if and only if $tr\big[\big(\textbf{I}+\frac{\textbf{A} \odot\textbf{A}}{1+k}\big)^{(k+1)}\big]=tr(\textbf{I})=(k+1)=0$.
\end{proof}

\par  To learn  the $\mathbb{G}$  over $\textbf{Z}=[\boldsymbol{\xi}^{(l)},\textbf{y}_s]$, we integrate  the above acyclicity constraint and the least-square loss between $\textbf{Z}$ and  $\textbf{ZA}$ into one model as follows.
\begin{equation}
\footnotesize
\begin{aligned}
&\ \quad \quad \quad \quad \quad \quad \min\limits_{\textbf{A}}\|\textbf{Z}-\textbf{Z}\textbf{A}\|^2 \\
&{\rm subject} \quad {\rm to} \quad tr\big[\big(\textbf{I}+\frac{\textbf{A} \odot\textbf{A}}{1+k}\big)^{(k+1)}\big]-(k+1)=0.
\label{noteras}
\end{aligned}
\end{equation}

%

\par We can first learn the low-dimensional feature representations $\boldsymbol{\xi}^{(l)}$. Based on the learnt $\boldsymbol{\xi}^{(l)}$, we obtain $\textbf{A}$ by solving the optimization problem given in Eq.(\ref{noteras}) following the process  in reference \cite{ZhengARX18}. That is,  after obtaining a stationary point $\textbf{A}$ of (\ref{noteras}), given a fixed threshold $\sigma >$  0, set any weights smaller than $\sigma$ in absolute value to zero. We obtain a new matric $\hat{\textbf{A}}$ by thresholding the edge weights of $\textbf{A}$ as follows.
\begin{equation}
\footnotesize
\begin{aligned}
\hat{\textbf{A}}[i][j] =\left\{
             \begin{array}{lcl}
             {0}  , &if \quad |\textbf{A}[i][j]| <\sigma, \\
             {\textbf{A}[i][j]}  ,&otherwise.
             \end{array}
        \right.
\label{Ahat}
\end{aligned}
\end{equation}
There exists a  directed edge from  variable $Z_i$ to variable $Z_j$ if $|\textbf{A}[i][j]| \geq\sigma$.

 \par Once $\hat{\textbf{A}}$ is learnt, we can identify the MB feature representations of class label through matrix $\hat{\textbf{A}}$, and  we can separate the low-dimensional representations $\boldsymbol{\xi}^{(l)}$ into two groups:  cause representations of the class variable $Y$, denoted as $\textbf{MB}(Y)$; and the other feature representations which are task-irrelevant (i.e. non-MB of $Y$), denoted as $\overline{\textbf{MB}}$$(Y)$, where $\boldsymbol{\xi}^{(l)}=\textbf{MB}(Y)\cup \overline{\textbf{MB}}(Y)$, and $\textbf{MB}(Y) \cap \overline{\textbf{MB}}(Y)=\emptyset$.

 \par Since the causal relationships between the class variable and its MB features are robust across different settings or environments, we can use $\textbf{MB}(Y)$ as the causal features to build the prediction model to be used for the target domain. This requires that the low-dimensional feature representations learnt by CAE need to be consistent with the underlying causal mechanism or relations represented by $\hat{\textbf{A}}$. However,  there is a difference between $\hat{\textbf{A}}$ and $\textbf{A}$, since  the edge weights of  $\textbf{A}$ that are smaller than $\sigma$ in absolute value  are set to zero. Hence, in learning low-dimensional feature representations, we  use the learnt  adjacency matrix $\hat{\textbf{A}}$  to constraint the low-dimensional representations as follows.


\begin{equation}
\footnotesize
\begin{aligned}
&\quad\quad\mathcal{L}_{C} = \|\textbf{Z}-\textbf{Z}\hat{\textbf{A}}\|^2.
\label{LossCSL}
\end{aligned}
\end{equation}

\subsection{Cross-entropy loss  $\mathcal{L}_{Y}$}

\par To improve the quality of predictions using the  causal feature representations, we incorporate the following cross-entropy loss into the objective function of CAE.
\begin{equation}
\footnotesize
\begin{aligned}
\mathcal{L}_{Y} = \ell(f(\textbf{MB}(Y)),\textbf{y}_{s}).
\label{LossY}
\end{aligned}
\end{equation}
where $\ell$($\cdot$) is a  cross-entropy loss, $f$ is a  classifier which uses the  causal feature representations. In this way, in the learning of the causal feature representations, CAE makes use of the label information of the samples in the source domain and aims to minimize  prediction error of a classifier using the causal feature representations.

\subsection{Optimization}
Based on  Eq.(\ref{DAE}), Eq.(\ref{CAER}), Eq.(\ref{LossCSL}) and Eq.(\ref{LossY}), the objective function of CAE in Eq.(\ref{LossCAE}) can be written as:
\begin{equation}
\footnotesize
\begin{aligned}
\mathcal{L}= &\frac{1}{2n}\|\textbf{X}_s-\hat{\textbf{X}}_s\|^2  + \lambda_1\|\textbf{Z}-\textbf{Z}\hat{\textbf{A}}\|^2 + \lambda_2\ell(f(\textbf{MB}(Y)),\textbf{y}_{s}) \\& + \lambda_3 \sum_{j=1}^{l}(||\textbf{W}_{1}^{(j)}||^2 + ||\textbf{b}_{1}^{(j)}||^2+||\textbf{W}_{2}^{(j)}||^2 + ||\textbf{b}_{2}^{(j)}||^2).
\label{WbSubPro}
\end{aligned}
\end{equation}


\par Now we show how to optimize the objective function. Note that the objective function involves several parameters, $\textbf{W}_{1}^{(j)}, \textbf{b}_{1}^{(j)}$, $\textbf{W}_{2}^{(j)}, \textbf{b}_{2}^{(j)}$ ($j=1,2,\cdots,l$) and $\hat{\textbf{A}}$. These  parameters are dependent on each other, thus the problem cannot be solved directly.  So we propose to  update $\textbf{W}_{1}^{(j)}, \textbf{b}_{1}^{(j)}$, $\textbf{W}_{2}^{(j)}, \textbf{b}_{2}^{(j)}$ ($j=1,2,\cdots,l$) and $\hat{\textbf{A}}$ in an alternate manner. First, we initialize $\hat{\textbf{A}}$ as the unit matrix, and update $\textbf{W}_{1}^{(j)}, \textbf{b}_{1}^{(j)}$, $\textbf{W}_{2}^{(j)}, \textbf{b}_{2}^{(j)}$. Since we implement CAE
using Tensorflow, once $\hat{\textbf{A}}$ is obtained, we use the Tensorflow framework to learn  $\textbf{W}_{1}^{(j)}, \textbf{b}_{1}^{(j)}$, $\textbf{W}_{2}^{(j)}, \textbf{b}_{2}^{(j)}$  by minimizing the objective function  in Eq.(\ref{WbSubPro}). Then, we  fix  $\textbf{W}_{1}^{(j)}, \textbf{b}_{1}^{(j)}$, $\textbf{W}_{2}^{(j)}, \textbf{b}_{2}^{(j)}$ and update $\hat{\textbf{A}}$.

 \par We learn $\hat{\textbf{A}}$ by  solving the problem in Eq.(\ref{noteras}). Once $\textbf{A}$ is learnt, we can obtain $\hat{\textbf{A}}$. We adopt the augmented Lagrangian method to solve the optimization problem given in Eq.(\ref{noteras}). That is, from Eq.(\ref{noteras}), we get the following  augmented Lagrangian.
\begin{equation}
\footnotesize
\begin{aligned}
\mathcal{L}(\textbf{Z},\alpha,\rho) = \|\textbf{Z}-\textbf{Z}\textbf{A}\|^2  + \alpha h(\textbf{A}) + \frac{\rho}{2} |h(\textbf{A})|^2.
\label{LossCSL2}
\end{aligned}
\end{equation}
where $h(\textbf{A}) = tr\big[\big(\textbf{I}+\frac{\textbf{A} \odot\textbf{A}}{1+k}\big)^{(k+1)}\big]-(k+1)$, $\alpha$  is the Lagrange multiplier, $\rho$ is is the penalty parameter.  We update $\textbf{A},\alpha,\rho$ by using the following rules.
\begin{equation}
\footnotesize
\begin{aligned}
&\textbf{A}^{(t+1)} = argmin \mathcal{L}(\textbf{Z},\alpha^{(t)},\rho^{(t)}),\\
&\rho^{(t+1)} =\left\{
             \begin{array}{lcl}
             { 10\rho^{(t)}}  , &if \quad |h(\textbf{A}^{(t+1)})| \geq \frac{1}{4}|h(\textbf{A}^{(t)})|, \\
             {\rho^{(t)} }  ,&otherwise.
             \end{array}
        \right. \\
&\alpha^{(t+1)} = \alpha^{(t)} + \rho^{(t)}h(\textbf{A}^{(t+1)}).
\label{SCLSubPro}
\end{aligned}
\end{equation}
where  $t$ represents the $t^{th}$ iteration, $\rho^{(1)}$ = 1 and  $\alpha^{(1)} = 0$ are the initial values of $\rho$ and $\alpha$.

\begin{algorithm}[t]
\footnotesize
\caption{The Causal AutoEncoder (CAE).}
\label{MyCAE}
\begin{algorithmic}[1]
\REQUIRE
Source domain data $\mathcal{D}_{s} = [\textbf{X}_{s},\textbf{y}_{s}]$\\ \quad \, \, Parameters $l$,  $k$, $\lambda_1$, $\lambda_2$, $\lambda_3$\\
\ENSURE
$\textbf{A}$, $\textbf{W}_{1}^{(j)}, \textbf{b}_{1}^{(j)}$, $\textbf{W}_{2}^{(j)}, \textbf{b}_{2}^{(j)}$, $j=1,2,\cdots,l$\\
\STATE Initialize parameters $\textbf{W}_{1}^{(j)}, \textbf{b}_{1}^{(j)}$, $\textbf{W}_{2}^{(j)}, \textbf{b}_{2}^{(j)}$\\
\STATE Initialize  $\hat{\textbf{A}}$: ($k$+1)*\textbf{I} identity matrix\\
\STATE Initialize the iteration variable $t\leftarrow0$
\STATE \textbf{Repeat}\\
\STATE \quad $t\leftarrow t+1$;\\
\STATE \quad  Fix $\hat{\textbf{A}}$, and update $\textbf{W}_{1}^{(j)}, \textbf{b}_{1}^{(j)}$, $\textbf{W}_{2}^{(j)}, \textbf{b}_{2}^{(j)}$ by Eq.(\ref{WbSubPro});  \\
\STATE \quad  Fix $\textbf{W}_{1}^{(j)}, \textbf{b}_{1}^{(j)}$, $\textbf{W}_{2}^{(j)}, \textbf{b}_{2}^{(j)}$, and  update $\textbf{A}$ by Eq.(\ref{SCLSubPro});\\
\STATE \quad  Obtain $\hat{\textbf{A}}$  by Eq.(\ref{Ahat});  \\
\STATE  \textbf{Until} $\mathcal{L}$ converges or max iteration is reached. \\
\end{algorithmic}
\end{algorithm}

\par The  proposed CAE algorithm is summarized in Algorithm \ref{MyCAE}. The default value of the maximum  iteration  is set to 10, and the default value used for determining convergence is set to 1E-8 in Algorithm \ref{MyCAE}.

\section{Experiments}

In this section, we evaluate the effectiveness of CAE by comparing it with several  feature selection and state-of-the-art  domain adaptation and stable learning methods.

%


\subsection{Experimental Settings}
\label{ExSet}
\subsubsection{\textbf{Datasets}}
\label{datasets}
The experiments are performed on three commonly used domain adaptation datasets, Office-Caltech10,  Amazon Review and Reuters-21578.

\par \textbf{Office-Caltech10} is a commonly used dataset for visual domain adaptation \cite{SunFS16}. The dataset  contains 2,533 images and 10 categories collected from four  real-world object domains: Caltech-256 (C), Amazon (A), Webcam (W) and DSLR (D).  We conduct experiments on this dataset with shallow features (SURF). The SURF features are encoded with 800-bin bag-of-words histograms.  We  evaluate all methods on the 12 cross-domain tasks: C$\rightarrow$A, C$\rightarrow$W, $\cdots$, D$\rightarrow$W.
\par \textbf{Amazon Review} is widely used as the benchmark dataset for cross-domain sentiment analysis. This dataset  contains a collection of product reviews  from Amazon.com about  four  types  of products: Books (B),  DVDS (D),  Electronics (E) and Kitchen appliances (K). There are about 1,000 positive reviews and 1,000 negative reviews in each product domain.   In our experiments, the preprocessed version of Amazon Review reported in \cite{WangCYHY19} is adopted. We construct 12 cross-domain tasks: B$\rightarrow$D, B$\rightarrow$E, $\cdots$, K$\rightarrow$E.
\par \textbf{Reuters-21578} contains a collection of  Reuters news documents that are organized with a hierarchical structure. It has three top   categories, Orgs (Or), People (Pe) and Places (Pl).  For fair comparison, the preprocessed version of Reuters-21578 reported in \cite{CaoLW18} is used. We select the 500 most frequent terms as features. From the dataset  we construct 3  cross-domain task: Or$\rightarrow$Pe, Or$\rightarrow$Pl, Pe$\rightarrow$Pl.

\tabcolsep 0.02in
\begin{table*}
\footnotesize
\centering
\caption{Accuracy (\%) of the 12 cross-domain tasks on the Office-Caltech10 dataset.}
\begin{tabular}{c|c||cccccccccccc||cc}
\hline
\multicolumn{2}{c||}{Methods}  &             C$\rightarrow$A      &C$\rightarrow$W &C$\rightarrow$D &A$\rightarrow$C  & A$\rightarrow$W  &A$\rightarrow$D  &W$\rightarrow$C  &W$\rightarrow$A &W$\rightarrow$D &D$\rightarrow$C &D$\rightarrow$A &D$\rightarrow$W &Avg &Avg rank\\
\hline\hline
\multirow{4}{*}{\begin{tabular}[c]{@{}c@{}}Feature\\ selection\end{tabular}}
&FCBF       &32.36 	 &26.10  &29.30  &29.03  &26.10   &22.29	&20.48  &20.15 	 &47.77  &19.59  &16.49  &33.90   &26.96 &10.75\\
&mRMR       &39.14 	 &29.15  &32.48  &26.09  &26.78   &28.66 	&19.32  &20.35	 &57.32  &22.26  &15.03  &38.64   &29.60 &10.75\\
&HINTON-MB  &40.71 	 &32.54  &43.94  &36.51  &30.51   &30.57 	&22.08  &23.28 	 &42.68  &19.77  &16.18  &23.73   &30.21 &8.63\\
&BAMB       &41.13 	 &32.54  &43.31  &36.60  &31.19   &31.85 	&21.82  &23.49 	 &45.86  &19.77  &16.18  &23.73   &30.62 &5.33\\\hline
\multirow{3}{*}{\begin{tabular}[c]{@{}c@{}}Domain\\ adaptation\end{tabular}}
&DTFC       &51.36 	 &36.94  &43.94  &40.16  &34.92   &\textbf{38.85} 	&29.11  &31.52 	 &77.07  &26.71  &25.99  &64.40   &41.80 &4.88\\
&TransNorm  &48.43    &37.50  &44.59  &40.54  &34.11   &37.36    &35.42  &37.01   &79.30  &31.42  &31.74  &74.22   &44.30 &3.83\\
&ALDA       &50.09    &\textbf{39.84}  &41.79  &35.93  &35.15   &33.20    &33.85  &38.47   &77.73  &32.81  &33.89  &73.44   &43.85 &3.92\\\hline
\multirow{4}{*}{\begin{tabular}[c]{@{}c@{}}Stable\\ learning\end{tabular}}
&CVS        &28.39 	 &31.19  &34.39  &30.63  &26.44   &24.84 	&19.67  &22.80 	 &49.68  &17.72  &16.70  &34.24   &28.05 &9.63\\
&CRLR       &48.75 	 &36.44  &39.81  &40.25  &35.93   &37.58 	&30.77   &33.87  &74.52  &27.28  &27.84   &60.29 &41.11    &5.25\\
&DGBR       &51.30 	 &38.50  &41.50  &42.09  &36.63   &35.19 	&36.19   &37.66  &\textbf{83.44}  &32.60  &32.51   &\textbf{81.00}   &45.72 &2.96\\
&DWR        &39.98 	 &28.81  &29.94  &35.08  &32.88   &26.11 	&26.27  &27.04 	 &62.74  &31.30  &31.83    &72.88    &37.07 &7.42\\\hline\hline
\multicolumn{2}{c||}{CAE}       &\textbf{52.36}	 &39.66 	&\textbf{45.54} &\textbf{43.46}  &\textbf{38.30} &35.66    &\textbf{37.40}  &\textbf{39.56}   &\textbf{83.44} &\textbf{33.39} &\textbf{34.86} &78.64 &\textbf{46.86} &\underline{1.46}\\
\hline
\end{tabular}
\label{accuracyOfOffice}
\end{table*}

\tabcolsep 0.02in
\begin{table*}
\footnotesize
\centering
\caption{Accuracy (\%) of the 12 cross-domain tasks on the Amazon dataset.}
\begin{tabular}{c|c||cccccccccccc||cc}
\hline
\multicolumn{2}{c||}{Methods}  &             B$\rightarrow$D       &B$\rightarrow$E &B$\rightarrow$K &D$\rightarrow$B  & D$\rightarrow$E  &D$\rightarrow$K  &E$\rightarrow$B  &E$\rightarrow$D &E$\rightarrow$K &K$\rightarrow$B &K$\rightarrow$D &K$\rightarrow$E &Avg &Avg rank\\
\hline\hline
\multirow{4}{*}{\begin{tabular}[c]{@{}c@{}}Feature\\ selection\end{tabular}}
&FCBF        &73.48	 &66.82  &67.53  &70.10  &72.92   &76.29 	&66.40  &68.83 	 &77.68  &67.50  &69.93  &77.02    &71.20  &9.25\\
&mRMR        &72.83 	 &68.77  &71.43  &73.15  &74.77   &77.19	&66.55  &68.38 	 &68.85  &68.85  &70.48  &78.22    &71.62  &8.04\\
&HINTON-MB   &72.79 	 &72.12  &73.74  &70.70  &71.92   &73.69 	&67.15  &69.23 	 &79.54  &69.15  &70.74  &78.37    &72.43  &8.08\\
&BAMB        &72.44 	 &71.67  &73.99  &71.30  &73.17   &75.24 	&66.00  &68.43 	 &79.79  &69.80  &71.94  &78.88    &72.72  &7.13\\\hline
\multirow{3}{*}{\begin{tabular}[c]{@{}c@{}}Domain\\ adaptation\end{tabular}}
&DTFC        &76.04 	 &76.92  &77.99  &\textbf{77.50}  &75.62   &76.54 	&72.85  &72.18 	 &79.58  &\textbf{72.75}  &71.24  &81.23    &75.87 &3.67\\
&TransNorm   &76.56   &76.70  &78.56  &68.01  &73.24   &75.19    &71.97  &72.60   &82.08  &68.55  &73.38  &82.17   &74.91  &4.42\\
&ALDA        &76.71   &75.09  &78.66  &67.77  &72.95   &76.66    &70.26  &73.39   &79.79  &69.43  &72.60  &80.37   &74.47  &4.71\\\hline
\multirow{4}{*}{\begin{tabular}[c]{@{}c@{}}Stable\\ learning\end{tabular}}
&CVS         &66.83 	 &66.47  &65.18  &65.80  &66.61   &65.38 	&62.60  &65.53   &69.43  &67.20  &70.87  &70.93    &66.90  &11.58\\
&CRLR        &77.94 	 &74.52  &75.08  &73.20  &72.17   &76.49 	&70.55  &73.49 	 &82.54  &69.60  &71.19  &81.08    &74.82  &4.42\\
&DGBR        &76.31 	 &73.27  &75.79  &66.60  &71.02   &74.06 	&71.17  &72.36 	 &79.97  &68.30  &72.46  &80.95    &73.52  &6.58\\
&DWR         &72.64 	 &71.97  &74.09  &70.05  &71.77   &74.43 	&66.70  &68.38 	 &78.10  &68.15  &70.54  &78.39    &72.10  &8.96\\\hline\hline
\multicolumn{2}{c||}{CAE}      &\textbf{79.15} 	 &\textbf{77.35}  &\textbf{78.83}  &73.62  &\textbf{76.82}   &\textbf{79.87} 	&\textbf{73.67}  &\textbf{74.18}	 &\textbf{83.96}  &71.20  &\textbf{75.44}  &\textbf{83.38}    &\textbf{77.29}  &\underline{1.17}\\
\hline
\end{tabular}
\label{accuracyOfAmazon}
\end{table*}


\tabcolsep 0.09in
\begin{table*}
\footnotesize
\centering
\caption{Accuracy (\%) of the 3 cross-domain tasks  on the Reuters-21578 dataset.}
\begin{tabular}{c|c||cccccc||cc}
\hline
\multicolumn{2}{c||}{Methods} &             Or$\rightarrow$Pe     &Or$\rightarrow$Pl &Pe$\rightarrow$Pl & Pe$\rightarrow$Or     &Pl$\rightarrow$Or &Pl$\rightarrow$Pe &Avg &Avg rank\\
\hline\hline
\multirow{4}{*}{\begin{tabular}[c]{@{}c@{}}Feature\\ selection\end{tabular}}
&FCBF         &67.21 	 &61.36  &56.63  &68.88 &63.68 &60.07 &62.97  &6.83  	\\
&mRMR         &73.10 	 &61.65  &54.78  &71.62 &62.70 &60.63 &64.08  &5.58  	 \\
&HINTON-MB    &62.00 	 &56.76  &55.15  &59.01 &59.15 &59.89 &58.66  &9.83   	 \\
&BAMB         &58.94 	 &61.55  &51.34  &53.92 &57.38 &59.70 &57.14  &10.5  	 \\\hline
\multirow{3}{*}{\begin{tabular}[c]{@{}c@{}}Domain\\ adaptation\end{tabular}}
&DTFC         &72.10           &62.60       &\textbf{60.53}       &73.16 &61.81 &60.72 &65.15       &4.00\\
&TransNorm    &72.66           &59.90       &57.55       &72.50 &62.59 &61.42 &64.44       &5.33\\
&ALDA         &70.94           &54.60       &58.76       &69.60 &63.57 &\textbf{62.69} &63.36       &5.75\\\hline
\multirow{4}{*}{\begin{tabular}[c]{@{}c@{}}Stable\\ learning\end{tabular}}
&CVS          &61.34 	 &60.21  &52.65  &64.29 &61.02 &59.24 &59.79  &9.75    \\
&CRLR         &73.92 	 &62.70  &57.66  &71.19 &64.37 &60.63 &65.07  &3.75  	 \\
&DGBR         &70.94 	 &62.99  &57.38  &72.64 &59.64 &58.15 &63.62  &6.75   	 \\
&DWR          &74.42 	 &58.19  &52.65  &67.34 &63.48 &58.68 &62.46  &7.92   	 \\\hline\hline
\multicolumn{2}{c||}{CAE}         &\textbf{75.66} 	 &\textbf{64.33}  &59.05 &\textbf{77.60} &\textbf{65.26} &60.33 &\textbf{67.03}  &\underline{2.00} \\
\hline\hline
\end{tabular}
\label{accuracyOfReuter}
\end{table*}

\subsubsection{Comparison Methods}
\label{comparedmethods}
 We  compare CAE with two correlation-based methods, FCBF \cite{YuL04} and mRMR \cite{PengLD05}, and two causal feature selection methods,  HITON-MB \cite{AliferisTS03} and BAMB \cite{BAMB}.  We also compare  CAE with three domain adaptation methods,  DTFC \cite{WeiFeature}, TransNorm \cite{WangJLWJ19} and ALDA \cite{LC20}, and four stable learning methods, CVS \cite{RCIT},  CRLR \cite{CRLR}, DGBR \cite{DGBR}  and  DWR \cite{KuangX0A020}. Note that the target domain data is unavailable in our problem setting, but the existing domain adaptation methods rely on the available target domain data to learn domain invariant representations. So in order to compare them with CAE in terms of generalizability to unknown target domains, using each of the existing domain adaptation methods,  we train a classifier on one pair of source and target domains, e.g. C$\rightarrow$A, then we apply the classifier to the task of adapting from the same source domain, but to a different target domain (which is unknown at training stage), to mimic the robust domain adaptation scenario as CAE has, e.g.  C$\rightarrow$W or C$\rightarrow$D.

\subsubsection{Implementation Details}
We implement CAE  using Tensorflow. In our experiments, the value of threshold $\sigma$  is set to 0.3, the number of  stacked layer $l$ is set to 2, the number of features $k$ is set to 50 on all datasets.  The initial values of $\alpha$ and $\rho$  are set to 0 and 1, respectively. On the Office-Caltech10 dataset, the  values of $\lambda_1$, $\lambda_2$ and $\lambda_3$  are set to 0.01, 1 and 1E-4, respectively.   On the Amazon Review dataset, the  values of $\lambda_1$, $\lambda_2$ and $\lambda_3$  are set to 10, 1 and  1E-4, respectively.   On the Reuters-21578 dataset, the  values of $\lambda_1$, $\lambda_2$ and $\lambda_3$  are set to 1, 0.1 and 1E-3, respectively.   For other comparison algorithms, we search for the optimal  hyper-parameter values and  report  the best results  in the experiments.

\par In the experiments, we utilize the classification accuracy on the  target domain data as the quality metric. For CAE, TransNorm, ALDA and DGBR,  each experiment is repeated 10 times, and we report the average performance.

\subsection{Experimental Results  and Analysis}
\subsubsection{Evaluation of classification performance}
The experimental results of CAE and its rivals on  the three datasets  are reported in Tables \ref{accuracyOfOffice} to \ref{accuracyOfReuter}. The best result in each task has been marked in  bold. From the experimental results, we have the following observations.
\par (a) Comparison with feature selection methods.
\begin{itemize}
\item  CAE performs better than FCBF and mRMR, which indicates the effectiveness of learning causal representations for robust domain adaptation.  The reasons  of this are twofold: on one hand,  FCBF and mRMR only select features that are  strongly relevant to class label, ignoring redundant features. However, in real cross-domain tasks, many document and image representations are sparse, redundant features can be used to enrich the knowledge of domains. On the other hand,  both of them may capture spurious correlations between features  and class variable that are not transferable across different target domains.

\item   CAE outperforms HINTON-MB and BAMB. HINTON-MB and BAMB  select causal features on the raw input data, making them  sensitive to the noise  in the data, whereas CAE reduces the impact of noise by mapping the input data into a low-dimensional feature space.  This indicates the advantages of mapping the input data into a low-dimensional space.

\item   HINTON-MB and BAMB are superior to  FCBF and mRMR in most cases, which indicates that the learnt causal features are more robust across different domains. However, both HINTON-MB and BAMB are worse than  FCBF and mRMR on the Reuters-21578 dataset. The reason of this is as follow: some true Markov blanket features may be discarded by HINTON-MB and BAMB, since Reuters-21578 is  a dataset that is organized with a hierarchical structure, meaning that learning MB of the class variable is more difficult if there are no sufficient samples.
\end{itemize}

\par (b) Comparison with domain adaptation methods.
\begin{itemize}
\item CAE performs better than DTFC, TransNorm and ALDA. These  domain adaptations methods use  the classifier trained on a given task to complete the other cross-domain tasks with the same source domain but an unknown target domain. Although the classifier may achieve good generalization performance on the given task,  the  classifier  achieves poor generalization performance on the other  tasks, due to the fact that the data distribution of other tasks is different from that of  the given task.  This shows that a normal domain adaption method cannot be used in the scenarios where the task-relevant and task-irrelevant target domain data is unavailable during the model training phase, since the classifier learnt for one target domain cannot be generalized to another domain.

\end{itemize}

 \par  (c) Comparison with stable learning methods.
\begin{itemize}
\item  CAE  outperforms CVS.  CVS also selects causal features from the raw input data, which is sensitive to the noise  in the data.  CVS uses a seed variable to separate causal and non-causal features and the seed variable has a great influence on the performance, whereas CAE learns  low-dimensional representations to reduces the impact of noise and does not require a seed variable as priori.

\item  CAE is superior to  DGBR and CRLR. This shows that  causal  representations learnt by CAE are more stable across domains.  DGBR and CRLR regard each variable as treatment variable and balance all of them together by global sample weighting.  For some treatment variables, global sample reweighting may not well balance the distributions of treated and control groups, making the estimated treatment effect imprecise and leading to poor performance.
\item  CAE achieves better performance than DWR. DWR  reduces correlation among  covariates through  sample reweighting, and the weights of many samples are assigned to 0,  which may cause  over-reduced sample size and lead to poor performance, while CAE makes full use of the knowledge in all samples.
\end{itemize}

\par In summary, CAE is superior to all state-of-the-art baseline methods. Therefore, we can conclude that the  cause representations learnt with CAE have good generalization performance on unknown target domain tata.  This indicates the effectiveness of CAE.

\begin{figure}
\centering
\subfigure[Office-Caltech10]{
    \begin{minipage}[b]{0.3\textwidth}
    \includegraphics[width=1\textwidth]{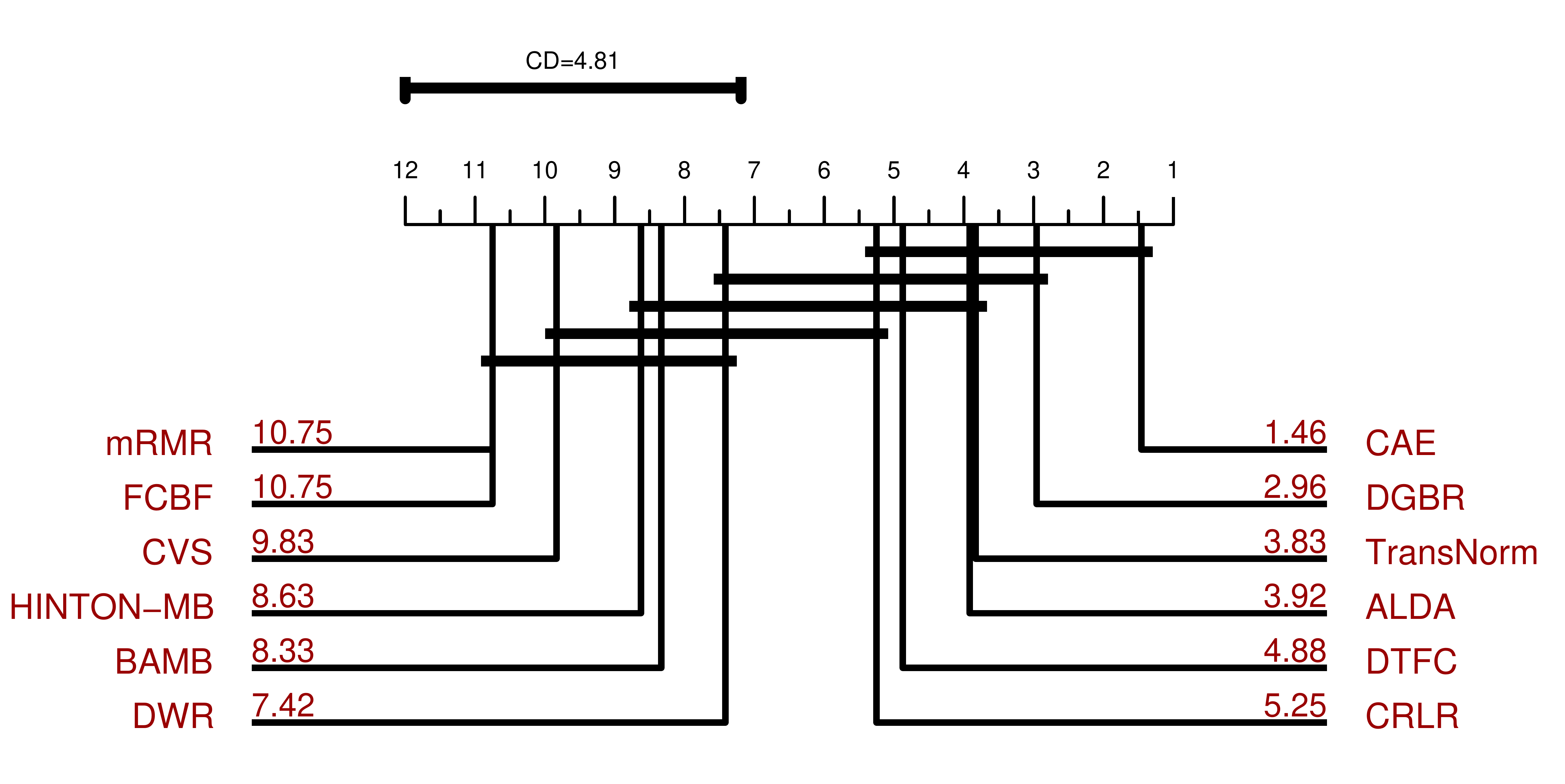}
    \end{minipage}
}
\subfigure[Amazon Review]{
    \begin{minipage}[b]{0.3\textwidth}
    \includegraphics[width=1\textwidth]{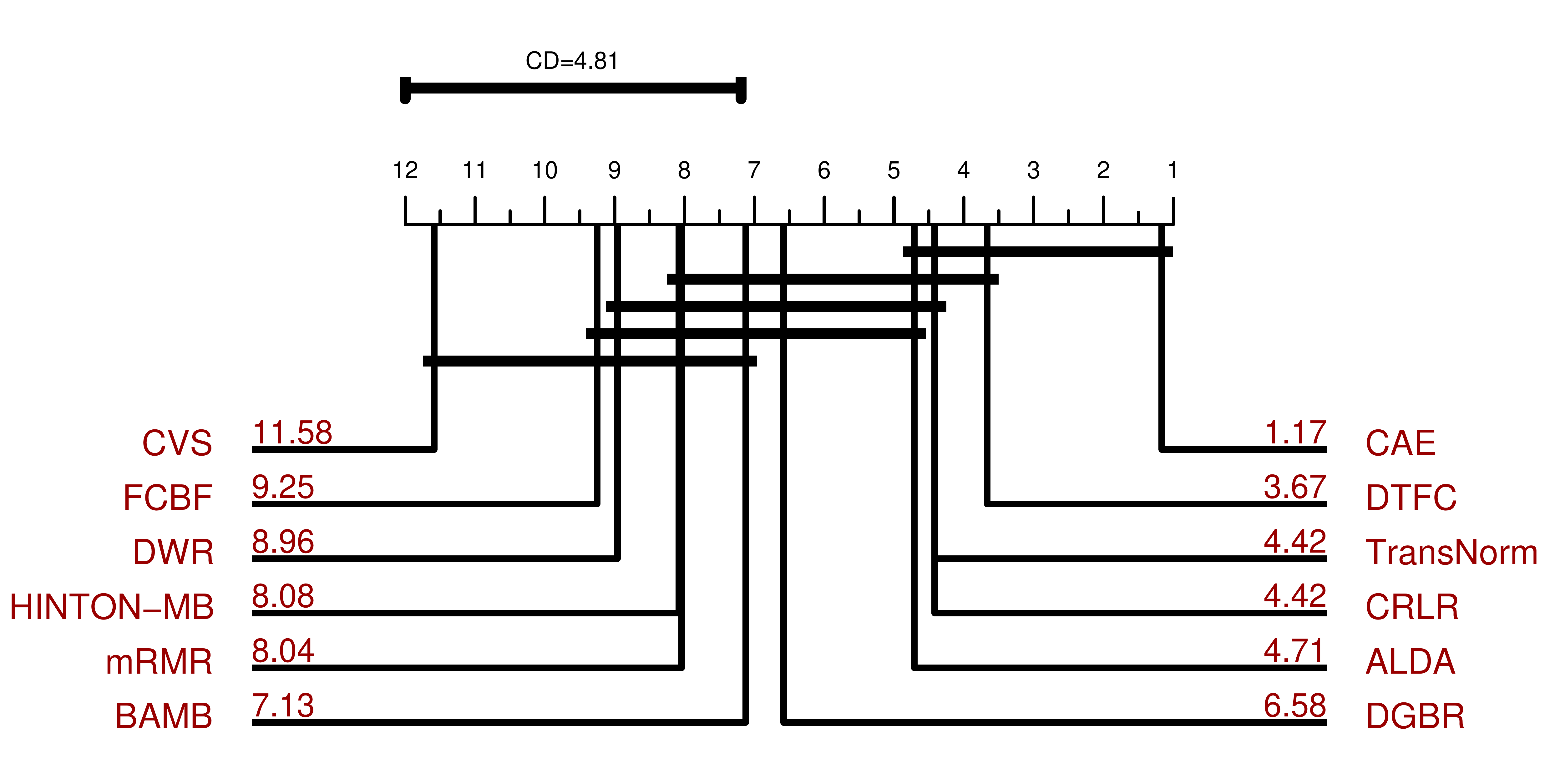}
    \end{minipage}
}
\subfigure[Reuters-21578]{
    \begin{minipage}[b]{0.3\textwidth}
    \includegraphics[width=1\textwidth]{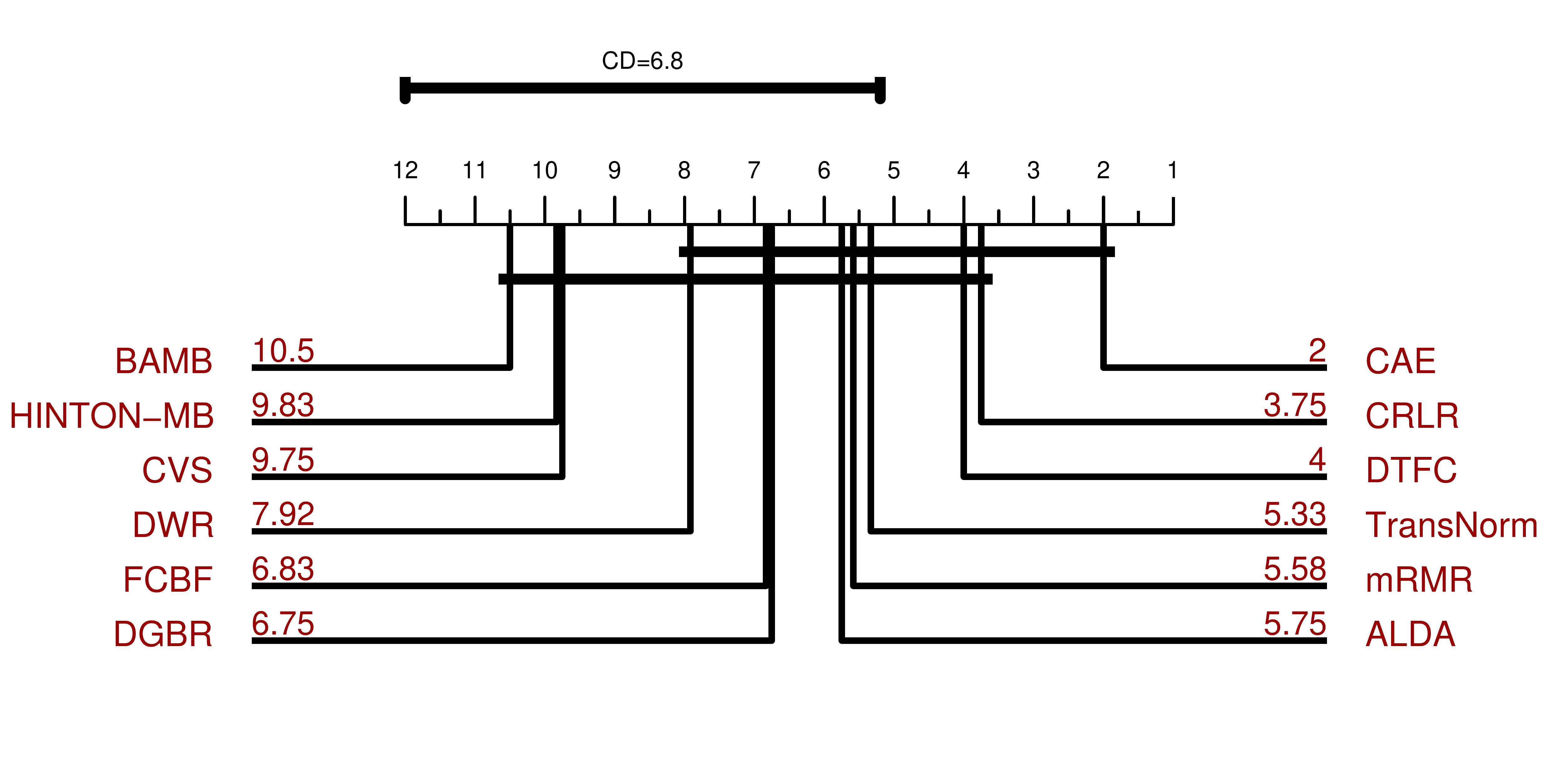}
    \end{minipage}
}
\caption{Comparison of CAE against the other algorithms with the Nemenyi test. The lower rank value is better (lower ranks to the right). }
\label{CD}
\end{figure}


\subsubsection{Statistical Tests}
To further compare the performance of CAE with that of its rivals,  the Friedman test  and  Nemenyi  test \cite{Demsar06}  are employed.

\par We first perform the Friedman test at the 5\% significance level under the null-hypothesis which states that all algorithms are performing equivalently (i.e., the average ranks of all algorithms are equivalent).  The  average ranks of CAE and its rivals are summarized in the last column of  Tables \ref{accuracyOfOffice} to \ref{accuracyOfReuter}.  It can be seen that the null hypothesis is rejected on all the three datasets. We also note that CAE performs better than its rivals (the lower rank value is better).
\par To further analyze the significant difference  between CAE and its rivals, we perform the Nemenyi test. Based on the test,  the  performances of two algorithms are significantly different if  the corresponding average ranks  differ by at least one critical difference (CD). The CD for the Nemenyi test \cite{Demsar06}  is calculated as follows.
\begin{equation*}
\footnotesize
\begin{aligned}
{\rm CD} = q_{a,m} \sqrt{\frac{m(m+1)}{6|\mathcal{D}|}}
\end{aligned}
\end{equation*}
where $\alpha$ is  the  significance level  and   $m$ is the number of comparison models, and $|\mathcal{D}|$ denotes the number of cross-domain tasks. In our experiments, $m$ = 12, $q_{a=0.05,m=12}$ = 3.268 at significance level $\alpha$ = 0.05. On the Office-Caltech10 and Amazon Review datasets, $|\mathcal{D}|$ = 12, and thus CD = 4.81. On the Reuters-21578 dataset, $|\mathcal{D}|$ = 6, and thus CD = 6.80.
\par Fig.\ref{CD} provides the  CD diagrams,  where the average rank of each  algorithm is marked along the axis (lower ranks to the right). On the Office-Caltech10 dataset,  CAE achieves comparable performance against DGBR, TransNorm, ALDA, DTFC and CRLR,  and CAE  significantly  outperforms the other algorithms. On the Amazon Review dataset, CAE  significantly  outperforms  CVS, FCBF, DWR, HINTON-MB, mRMR and BAMB, and CAE achieves comparable performance against the other algorithms. On the Reuters-21578 dataset,  CAE  significantly  outperforms  BAMB, HINTON-MB, CVS, DWR, FCBF and DGBR, and CAE achieves comparable performance against the other algorithms.  CAE is the only algorithm that achieves the lowest ranking values  in all three datasets.

\subsection{Ablation and Sensitivity Analysis}
\label{AnaDRAE}
To study the effect of different components of the objective function of CAE, we perform an ablation study on nine cross-domain tasks. To investigate the effect of the causal structure loss $\mathcal{L}_{C}$,  we remove $\mathcal{L}_{C}$ from Eq.(\ref{LossCAE}) and keep other components  unchanged, and we denote this version of CAE  as ``CAE w/o $\mathcal{L}_{C}$''. To investigate the effect of the cross-entropy loss $\mathcal{L}_{Y}$, similarly we use  the ``CAE w/o $\mathcal{L}_{Y}$'', for which the $\mathcal{L}_{Y}$  component is removed from Eq.(\ref{LossCAE}) while the other components remain unchanged. The experimental results are summarized in Table \ref{abOffice} to \ref{abReuter}. We observe that ``CAE w/o $\mathcal{L}_{C}$'' is inferior to CAE, which indicates the effectiveness of  the  $\mathcal{L}_{C}$  component. On the tasks of A$\rightarrow$D and D$\rightarrow$C,  ``CAE w/o $\mathcal{L}_{C}$'' achieves better results than CAE, but ``CAE w/o $\mathcal{L}_{C}$'' is inferior to CAE on the tasks of A$\rightarrow$C and D$\rightarrow$A, D$\rightarrow$W.  The reason is that  ``CAE w/o $\mathcal{L}_{C}$'' may capture spurious correlations between features and the class variable, as it does not use the $\mathcal{L}_{C}$ term  to separate the learnt low-dimensional representations into two groups,  the spurious correlations may facilitate the learning of the tasks of A$\rightarrow$D, W$\rightarrow$C and D$\rightarrow$C, but they may not be transferable across different target domains.   Comparing  the  result  of  ``CAE w/o $\mathcal{L}_{Y}$''  and  CAE, we  note that the performance of ``CAE w/o $\mathcal{L}_{Y}$'' drops, which shows the necessity of  incorporating the label information of source domain to improve the quality of  causal   representations for predictions. We can conclude that both causal structure loss $\mathcal{L}_{C}$ and cross-entropy loss $\mathcal{L}_{Y}$ play a major role in  CAE.

\tabcolsep 0.02in
\begin{table*}
\footnotesize
\centering
\caption{Ablation study on the  Amazon Review dataset.}\label{abOffice}
\begin{tabular}{c||cccccccccccc||c}
\hline
Methods  &             C$\rightarrow$A      &C$\rightarrow$W &C$\rightarrow$D &A$\rightarrow$C  & A$\rightarrow$W  &A$\rightarrow$D  &W$\rightarrow$C  &W$\rightarrow$A &W$\rightarrow$D &D$\rightarrow$C &D$\rightarrow$A &D$\rightarrow$W &Avg \\
\hline\hline
CAE w/o $\mathcal{L}_{C}$       &50.96 	 &38.64  &43.82  &41.76  &36.20   &\textbf{38.73}	&37.31  &38.18 	 &80.89  &\textbf{33.74}  &33.79  &76.54  &45.88 \\
CAE w/o $\mathcal{L}_{Y}$       &45.72 	 &37.96  &44.59  &35.26  &32.20   &32.48 	&30.45  &35.38	 &73.24  &32.94  &34.24  &71.52   &42.17 \\\hline\hline
CAE       &\textbf{52.36}	 &\textbf{39.66} 	&\textbf{45.54} &\textbf{43.46}  &\textbf{38.30} &35.66    &\textbf{37.40}  &\textbf{39.56}   &\textbf{83.44} &33.39 &\textbf{34.86} &\textbf{78.64} &\textbf{46.86} \\
\hline
\end{tabular}
\end{table*}

\tabcolsep 0.03in
\begin{table*}
\footnotesize
\centering
\caption{Ablation study on  the Office-Caltech10 dataset.}\label{abAmazon}
\begin{tabular}{c||cccccccccccc||c}
\hline
Methods  &              B$\rightarrow$D       &B$\rightarrow$E &B$\rightarrow$K &D$\rightarrow$B  & D$\rightarrow$E  &D$\rightarrow$K  &E$\rightarrow$B  &E$\rightarrow$D &E$\rightarrow$K &K$\rightarrow$B &K$\rightarrow$D &K$\rightarrow$E  &Avg \\
\hline\hline
CAE w/o $\mathcal{L}_{C}$       &\textbf{79.15} 	 &\textbf{77.47}  &77.74  &73.00  &75.57   &78.98	&72.90  &73.08 	 &83.39  &68.60  &74.69  &81.93  &76.37 \\
CAE w/o $\mathcal{L}_{Y}$       &78.69 	 &77.03  &78.09  &70.90  &75.72   &78.98	&70.35  &68.08	 &79.29  &69.00  &70.54  &77.18   &74.49 \\\hline\hline
CAE       &\textbf{79.15} 	 &77.35  &\textbf{78.83}  &\textbf{73.62 } &\textbf{76.82}   &\textbf{79.87} 	&\textbf{73.67}  &\textbf{74.18}	 &\textbf{83.96}  &\textbf{71.20}  &\textbf{75.44}  &\textbf{83.38}    &\textbf{77.29} \\
\hline
\end{tabular}
\end{table*}

\tabcolsep 0.12in
\begin{table*}
\footnotesize
\centering
\caption{Ablation study on  the Reuters-21578 dataset.}\label{abReuter}
\begin{tabular}{c||cccccc||c}
\hline
Methods  &              Or$\rightarrow$Pe     &Or$\rightarrow$Pl &Pe$\rightarrow$Pl & Pe$\rightarrow$Or     &Pl$\rightarrow$Or &Pl$\rightarrow$Pe  &Avg \\
\hline\hline
CAE w/o $\mathcal{L}_{C}$       &73.34 	 &62.99  &58.12  &74.55  &63.87   &\textbf{60.33}	 &65.53 \\
CAE w/o $\mathcal{L}_{Y}$       &72.93 	 &61.45  &58.49  &71.79  &63.39   &58.69 	  &64.46 \\\hline\hline
CAE       &\textbf{75.66} 	 &\textbf{64.33}  &\textbf{59.05} &\textbf{77.60} &\textbf{65.26} &\textbf{60.33} &\textbf{67.03} \\
\hline
\end{tabular}
\end{table*}

\begin{figure*}
\centering
\subfigure[Parameter $\lambda_1$]{
    \begin{minipage}[b]{0.28\textwidth}
    \includegraphics[width=1\textwidth]{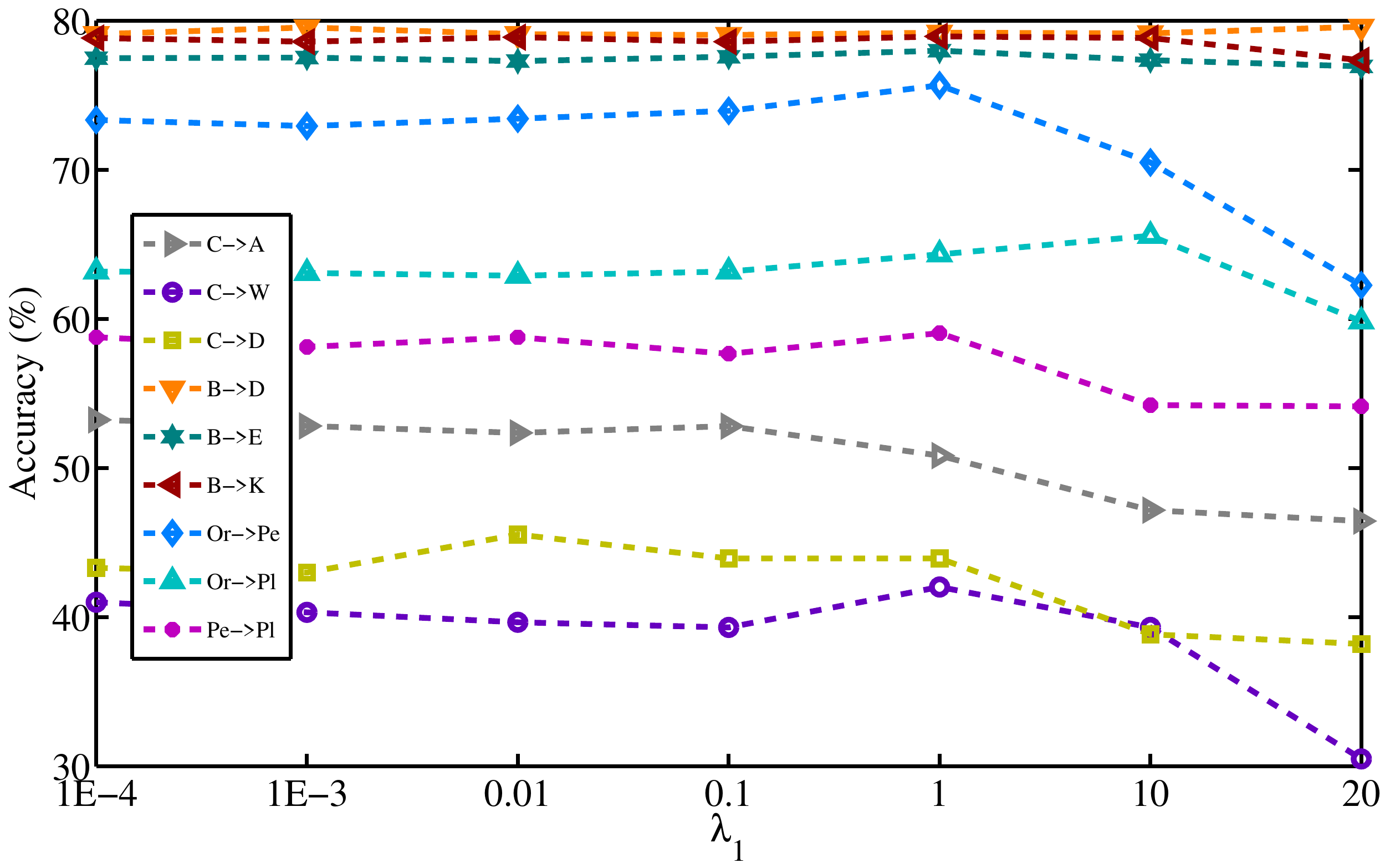}
    \end{minipage}
}
\subfigure[Parameter $\lambda_2$]{
    \begin{minipage}[b]{0.28\textwidth}
    \includegraphics[width=1\textwidth]{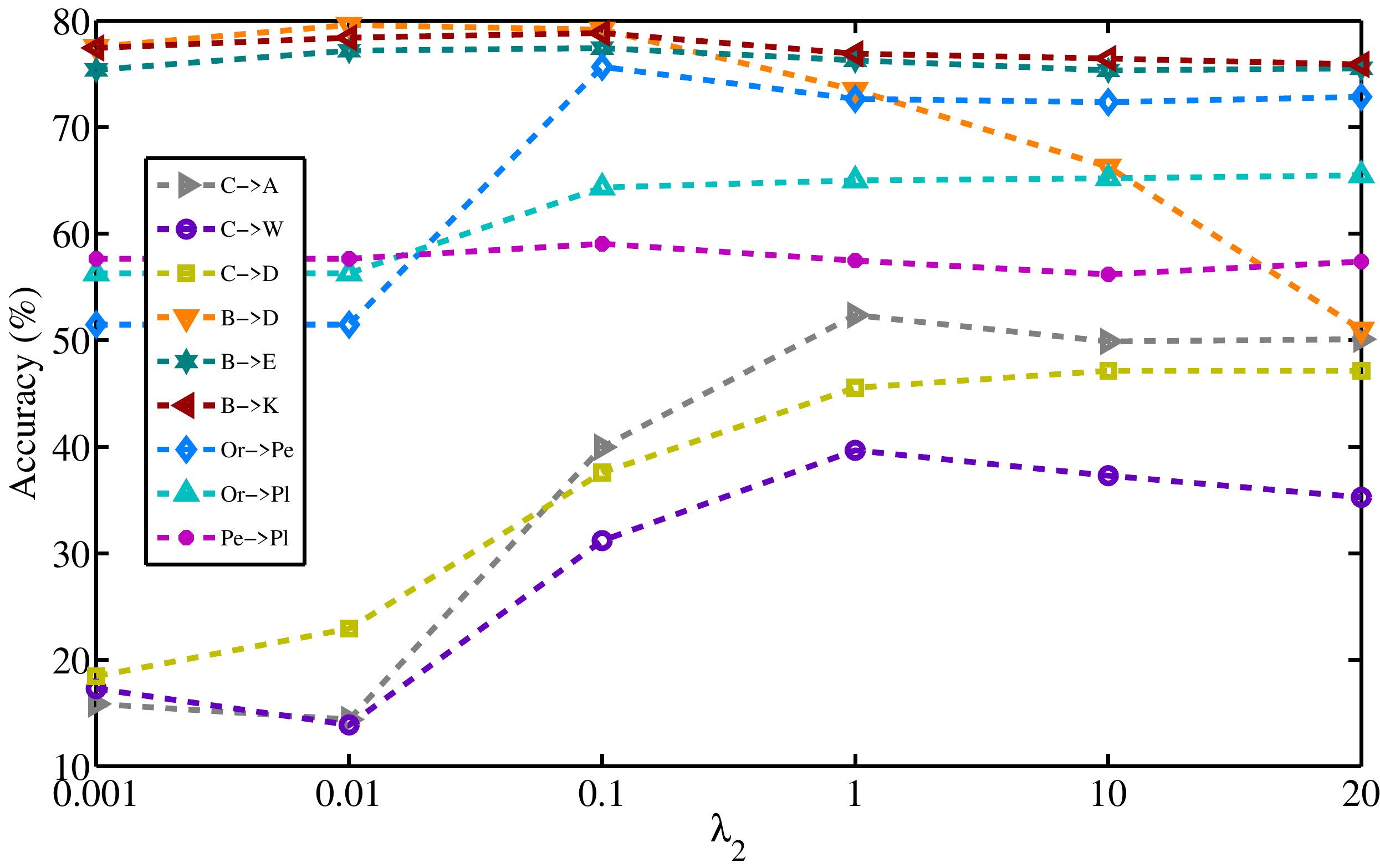}
    \end{minipage}
}
\subfigure[Parameter $\lambda_3$]{
    \begin{minipage}[b]{0.28\textwidth}
    \includegraphics[width=1\textwidth]{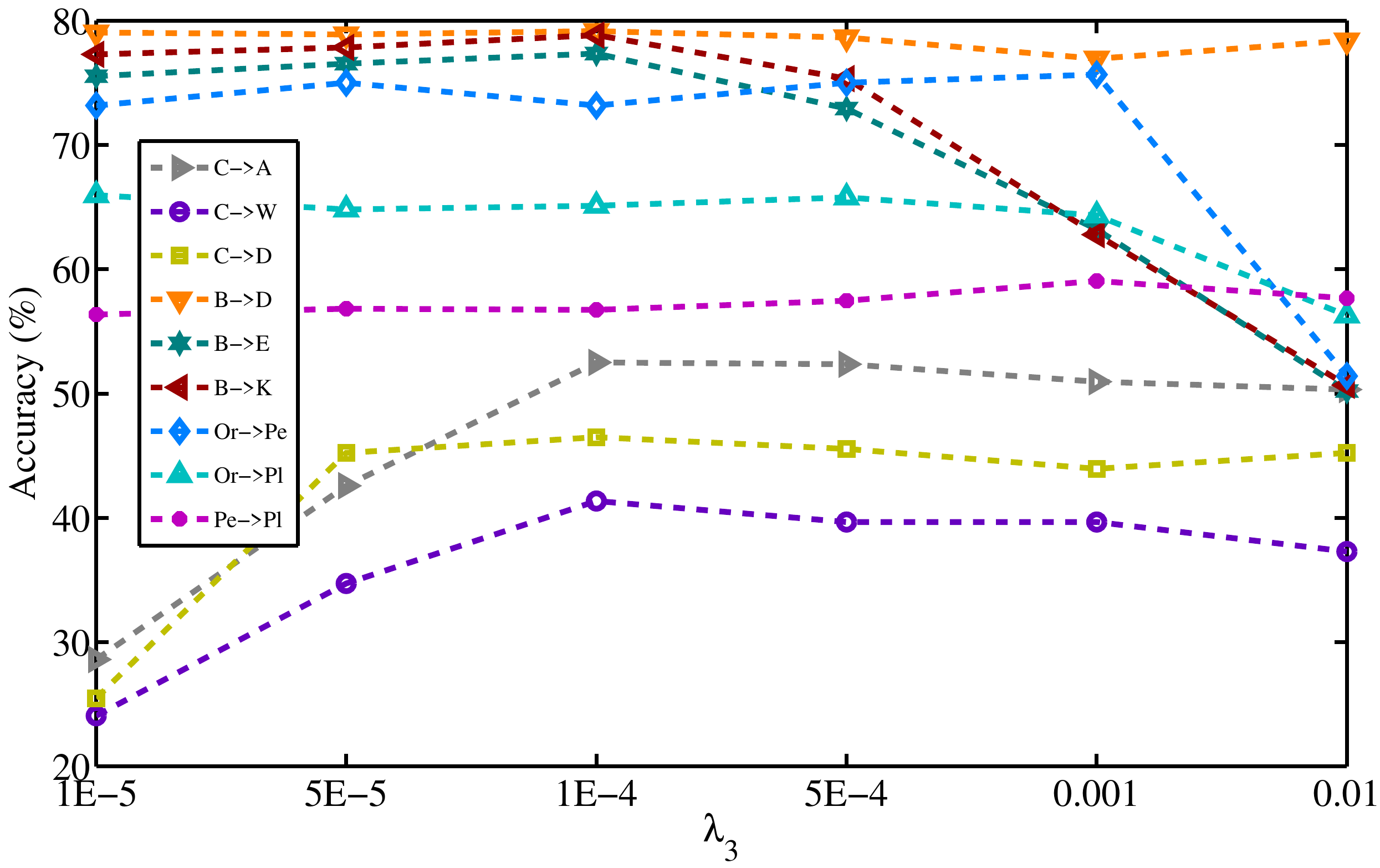}
    \end{minipage}
}
\subfigure[Parameter $l$]{
    \begin{minipage}[b]{0.28\textwidth}
    \includegraphics[width=1\textwidth]{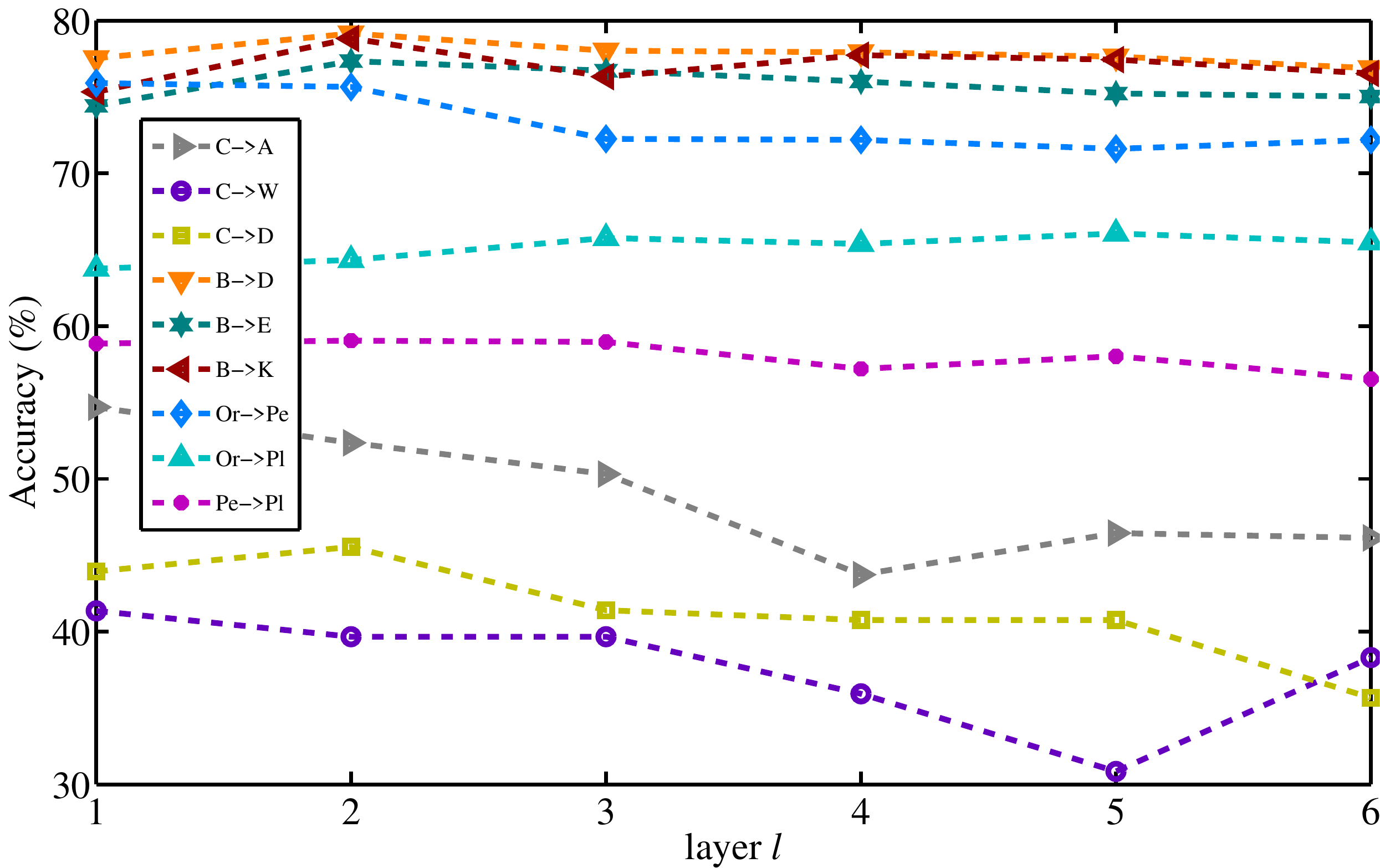}
    \end{minipage}
}
\subfigure[Parameter $k$]{
    \begin{minipage}[b]{0.28\textwidth}
    \includegraphics[width=1\textwidth]{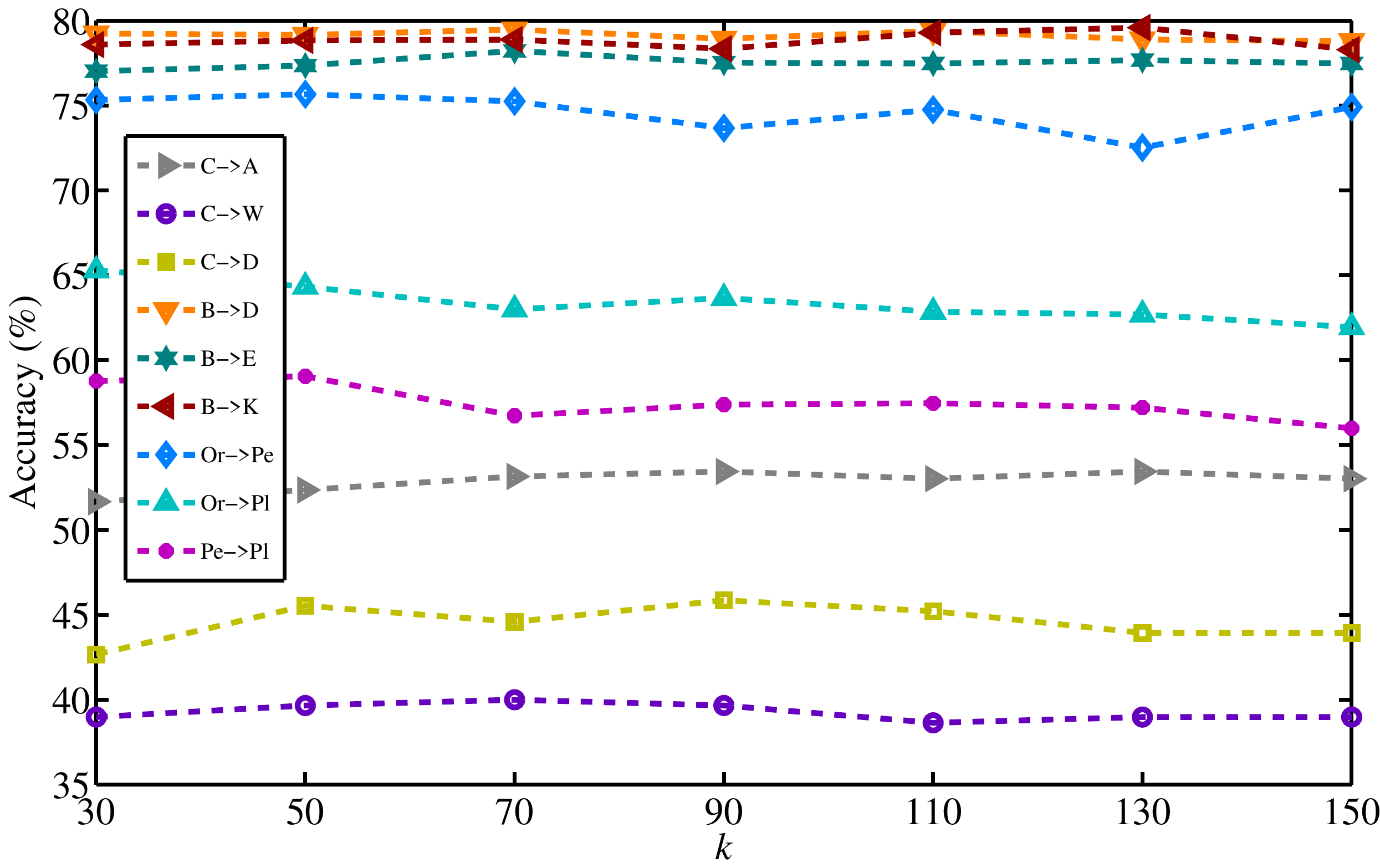}
    \end{minipage}
}
\caption{Parameter sensitivity study for CAE on  randomly selected cross-domain tasks. }
\label{parameteranalysis}
\end{figure*}

%

\par We also investigate the sensitivity of CAE with respect to parameters.   CAE  has five parameters,  the number of stacked layers $l$, the number of low-dimensional features $k$,  hyper-parameters  $\lambda_1$, $\lambda_2$ and $\lambda_3$.  When tuning a specific parameter, the other  parameters remain unchanged.

\par \textbf{Parameters} \textbf{$\lambda_1$}, \textbf{$\lambda_2$} and \textbf{$\lambda_3$}. The classification accuracies with different  values for $\lambda_1$, $\lambda_2$ and $\lambda_3$ are plotted in Fig.\ref{parameteranalysis}(a)(b)(c).  We note that CAE is  sensitive to  the three hyper-parameters. For  $\lambda_1$, [1,10] and   [0.01,1] are the optimal value ranges for Amazon Review and the other datasets, respectively. For $\lambda_2$,  [0.1,1] is the optimal  value range for all datasets. For $\lambda_3$,  [5E-4,0.001] and [5E-5,1E-4] are the optimal  value range for Reuters-21578 and the other datasets, respectively.  The values of $\lambda_1$  and $\lambda_2$  should be neither too small nor too large. If the value of each of them is too large, the effect of other three components in Eq.(\ref{LossCAE}) will be weakened.  If the value of $\lambda_1$ is too small,   the low-dimensional representations learnt with a deep autoencoder model cannot well split into two groups, i.e. causal representations and task-irrelevant representations. Since autoencoder is an unsupervised model, if the value of $\lambda_2$ is too small,  the low-dimensional representations learnt with a deep autoencoder model may contain less  causal representations and contain more task-irrelevant representations. This will lead to poor performance.
\par \textbf{Parameters} \textbf{$l$} and \textbf{$k$}. We plot the classification accuracies with  different values for $l$  and $k$ in Fig.\ref{parameteranalysis}(d)(e).  We observe that CAE is  sensitive to  the two hyper-parameters.  For $l$,  [1,3] is the optimal  value range for all datasets. If the value of $l$ is too large,  it will make the  training process  difficult and it is easy to cause  over fitting,  leading to performance degradation.  The value of $k$ cannot be set too large. For $k$,  [50,90] is the optimal  value range for all datasets. If the value of  $k$ continues to increase, the training time increases a lot but the
performance drops.

\section{Conclusion and Future Studies}
\label{conclusion}
In this paper, we  study a new problem, robust domain adaptation,  in which the unlabeled target domain data is unknown during the model  training phase.   To tackle this problem, we propose the Causal AutoEncoder (CAE) algorithm, which jointly optimizes a deep autoencoder model and a causal structure learning model to extract   casual  representations only using data from a single source domain. To the best of our knowledge, CAE is the first method to directly learn  causal representations using a deep autoencoder model. Our method  provides a new insight for learning  causal representation using deep neural networks. Extensive experiments are conducted on three real-world  datasets, and the experimental results demonstrate the effectiveness of our proposed method. Future work could study how to combine autoencoder and other deep neural networks to learn higher quality  causal representations for robust domain adaptation.

\section*{Acknowledgments}
This work is supported by the National Key Research and Development Program of China (under grant 2020AAA0106100), in part by the National Natural Science Foundation of China (under Grant 61876206), and in part by the Open Project Foundation of Intelligent Information Processing Key Laboratory of Shanxi Province (under grant CICIP2020003).

\section*{Appendix}
\begin{appendix}
In this section, we will give a brief introduction to Markov Blanket (MB) \cite{neufeld1993pearl}. Let $\emph{\textbf{U}}$ denote the set of random variables. $\mathbb{P}$   represents a joint probability distribution over $\emph{\textbf{U}}$, and $\mathbb{G}$   is a directed acyclic graph (DAG) over $\emph{\textbf{U}}$. In a  DAG, if there exists a directed edge from variable $\emph{X}$ to variable $\emph{Y}$, then $\emph{X}$ is a parent of  $\emph{Y}$ and $\emph{Y}$ is a child of $\emph{X}$.   $\emph{X}$ is an ancestor of $\emph{Y}$ (i.e., non-descendant of $\emph{Y}$) and $\emph{Y}$ is a descendant of $\emph{X}$ if there exists a directed path from $\emph{X}$ to $\emph{Y}$. The triplet $<\emph{\textbf{U}}, \mathbb{G} , \mathbb{P} >$  constitutes  a Bayesian Network (BN) if and only if $<\emph{\textbf{U}}, \mathbb{G} , \mathbb{P}>$ satisfies the Markov condition \cite{CPS}: each variable is conditionally independent of  variables in its non-descendants given its parents in $\mathbb{G}$. In the following, we will first give the definitions of conditional independence and faithfulness, then  briefly introduce MB.

\begin{definition}[Conditional Independence \cite{neufeld1993pearl}]\label{def1}
Given a conditioning set \textbf{Z},  variable X is conditionally independent of  variable Y  if and only if $P(X|Y,\textbf{Z}) = P(X|\textbf{Z})$.
\end{definition}

\begin{definition}[Faithfulness \cite{CPS}]\label{def4}
Given a Bayesian Network $<\textbf{U},\mathbb{G},\mathbb{P}>$,  $\mathbb{G}$ is faithful to $\mathbb{P}$ if and only if all the conditional independencies appear in $\mathbb{P}$ are entailed by $\mathbb{G}$.  $\mathbb{P}$ is faithful if and only if there is a DAG $\mathbb{G}$ such that $\mathbb{G}$ is faithful to $\mathbb{P}$.
\end{definition}

\begin{figure}
  \centering
  \includegraphics[width=6.5cm]{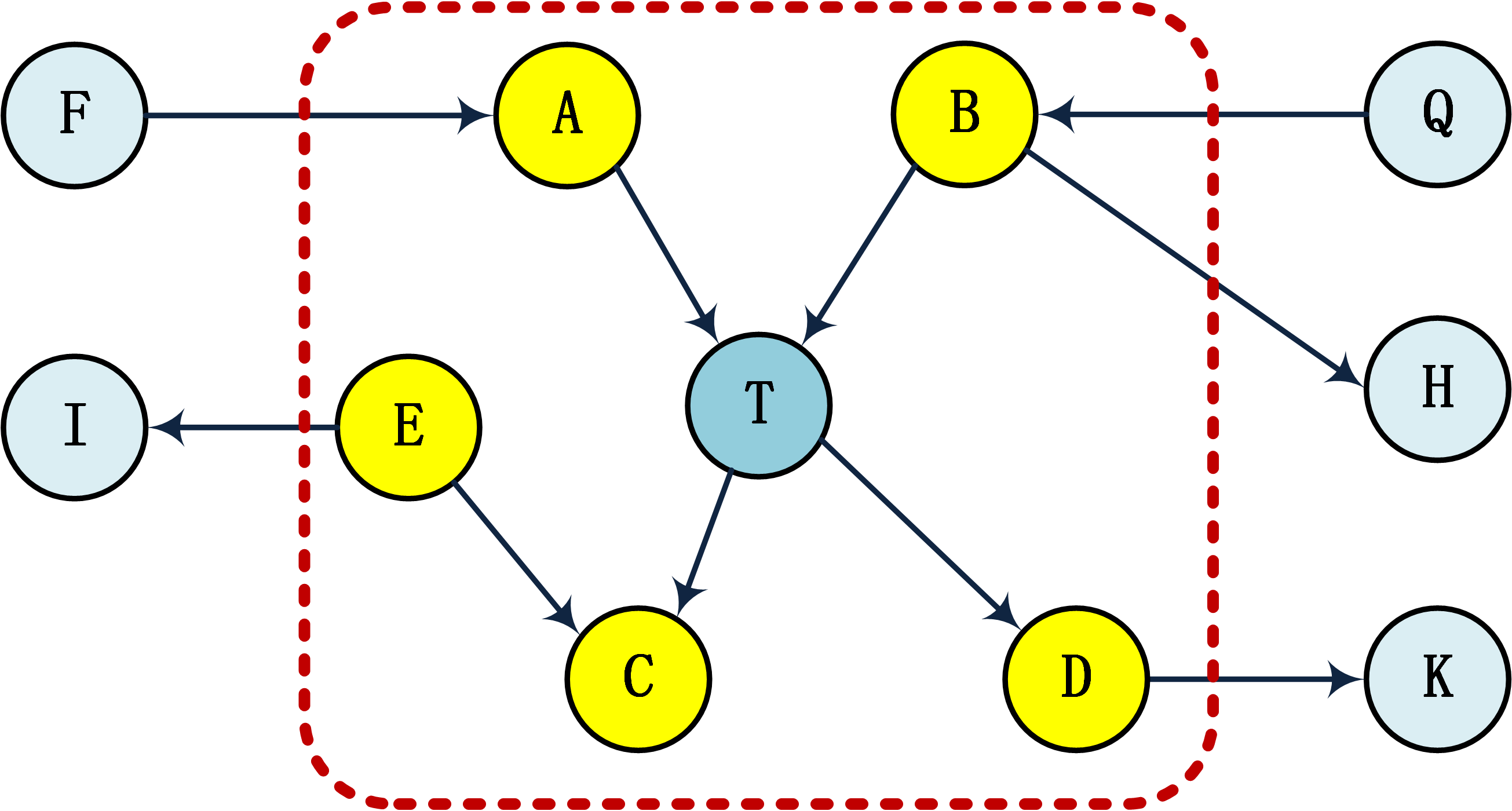}
  \caption{ An example of an MB of variable \emph{T} (yellow variables) in a Bayesian network.}
  \label{MB}
\end{figure}

The notion of a Markov blanket (MB)  was proposed by  Pearl et al. \cite{neufeld1993pearl} in the context of a Bayesian network (BN).  If  the faithfulness assumption (see Definition \ref{def4}) holds in a BN, the \textbf{MB} of a target variable \emph{T} consists of parents, children and spouses (other parents of the children) of \emph{T} and  is uniqueness, is denoted as \textbf{MB}$_{T}$ (Definition \ref{def5}). Fig. \ref{MB} gives an example of a MB in a BN.  The MB of \emph{T} includes \emph{A} and \emph{B} (parents), \emph{C} and \emph{D} (children), and \emph{E} (spouse), that is, \textbf{MB}$_{T}$ = \{\emph{A}, \emph{B}, \emph{C}, \emph{D}, \emph{E}\}. The MB of a target variable is the minimal feature subset that renders all other features conditionally independent of the target (see Theorem \ref{thmmb}). As shown in Fig.\ref{MB},  given conditioning set \textbf{MB}$_{T}$,   variables \emph{F}, \emph{I}, \emph{Q}, \emph{H} and \emph{K}  all are conditionally independent of the target  \emph{T}. The MB of the class variable has been proved to be the optimal feature subset for feature selection under the faithfulness condition \cite{TsamardinosA03}. As a concept from  Bayesian network (BN), the MB of class variable  provides the local causal structure around the class variable and thus uncovers the causal relationships between features and the class variable \cite{neufeld1993pearl}.  Compared with  non-causal feature selection  considering  exploit statistical associations or dependences between features and the class variable, causal feature selection uncovers the underlying mechanism of the occurrence of the class variable that is  persistent across different settings or environments,  and thus it has potential abilities to select more robust and interpretable features than non-causal feature selection methods in non-static environment. Recently, due to the interpretability and robustness, MB has received increasing attention and has been widely applied for causal feature selection \cite{yu2020causality}.

\begin{definition}[Markov Blanket \cite{neufeld1993pearl}]\label{def5}
In a faithful BN, the MB of a target variable T is denoted as \textbf{MB}$_{T}$, which consists of parents, children and spouses (other parents of the children) of T and  is uniqueness.
\end{definition}

\begin{theoremNoParens}[\cite{neufeld1993pearl}]\label{thmmb}
Given the Markov blanket (\textbf{MB}$_{T}$) of a target variable T, all other variables  in \textbf{U} $\setminus$\textbf{MB}$_{T}$$\setminus$\{T\} are conditionally independent of T, that is, X $\!\perp\!\!\!\perp$ T $|$ \textbf{MB}$_{T}$, for $\forall$ X $\subseteq$  \textbf{U} $\setminus$\textbf{MB}$_{T}$$\setminus$\{T\} .
\end{theoremNoParens}

%

  \end{appendix}

\section*{REFERENCES}
\bibliographystyle{elsarticle-num}
\bibliography{CAEref}

\begin{thebibliography}{10}
\expandafter\ifx\csname url\endcsname\relax
  \def\url#1{\texttt{#1}}\fi
\expandafter\ifx\csname urlprefix\endcsname\relax\def\urlprefix{URL }\fi
\expandafter\ifx\csname href\endcsname\relax
  \def\href#1#2{#2} \def\path#1{#1}\fi

\bibitem{zhuangsurvey}
F.~Zhuang, Z.~Qi, K.~Duan, D.~Xi, Y.~Zhu, H.~Zhu, H.~Xiong, Q.~He, A
  comprehensive survey on transfer learning, CoRR abs/1911.02685.

\bibitem{RenXY20}
C.~Ren, X.~Xu, H.~Yan, Generalized conditional domain adaptation: {A} causal
  perspective with low-rank translators, {IEEE} transactions on cybernetics
  50~(2) (2020) 821--834.

\bibitem{ChenSLYW19}
Y.~Chen, S.~Song, S.~Li, L.~Yang, C.~Wu, Domain space transfer extreme learning
  machine for domain adaptation, {IEEE} transactions on cybernetics 49~(5)
  (2019) 1909--1922.

\bibitem{WeiK19MIMT}
P.~Wei, Y.~Ke, Knowledge transfer based on multiple manifolds assumption, in:
  Proceedings of the 28th {ACM} International Conference on Information and
  Knowledge Management, November 3-7, Beijing, China, 2019, pp. 279--287.

\bibitem{LC20}
M.~Chen, S.~Zhao, H.~Liu, D.~Cai, Adversarial-learned loss for domain
  adaptation, in: Proceedings of the Thirty-Fourth {AAAI} Conference on
  Artificial Intelligence, February 7-12, New York, USA, 2020, pp. 3521--3528.

\bibitem{DengLZ19}
Z.~Deng, Y.~Luo, J.~Zhu, Cluster alignment with a teacher for unsupervised
  domain adaptation, in: 2019 {IEEE/CVF} International Conference on Computer
  Vision, {ICCV} ,October 27 - November 2, Seoul, Korea (South),, 2019, pp.
  9943--9952.

\bibitem{LXSDH19}
S.~Li, C.~H. Liu, B.~Xie, L.~Su, Z.~Ding, G.~Huang, Joint adversarial domain
  adaptation, in: Proceedings of the 27th {ACM} International Conference on
  Multimedia, October 21-25, Nice, France, 2019, pp. 729--737.

\bibitem{Kang0YH19}
G.~Kang, L.~Jiang, Y.~Yang, A.~G. Hauptmann, Contrastive adaptation network for
  unsupervised domain adaptation, in: {IEEE} Conference on Computer Vision and
  Pattern Recognition, June 16-20, CA, USA, 2019, pp. 4893--4902.

\bibitem{LongC0J18}
M.~Long, Z.~Cao, J.~Wang, M.~I. Jordan, Conditional adversarial domain
  adaptation, in: Advances in Neural Information Processing Systems 31,
  December 3-8, Canada, 2018, pp. 1647--1657.

\bibitem{kunZhang2020}
K.~Zhang, M.~Gong, P.~Stojanov, B.~Huang, C.~Glymour,
  \href{https://arxiv.org/abs/2002.03278}{Domain adaptation as a problem of
  inference on graphical models}, CoRR abs/2002.03278.
\newline\urlprefix\url{https://arxiv.org/abs/2002.03278}

\bibitem{prop}
T.~Teshima, I.~Sato, M.~Sugiyama, Few-shot domain adaptation by causal
  mechanism transfer, CoRR abs/2002.03497.

\bibitem{WangJ19}
J.~Wang, J.~Jiang, Conditional coupled generative adversarial networks for
  zero-shot domain adaptation, in: {IEEE/CVF} International Conference on
  Computer Vision, {ICCV}, October 27 - November 2,Seoul, Korea (South),,
  {IEEE}, 2019, pp. 3374--3383.

\bibitem{ZSDA}
K.~Peng, Z.~Wu, J.~Ernst, Zero-shot deep domain adaptation, in: V.~Ferrari,
  M.~Hebert, C.~Sminchisescu, Y.~Weiss (Eds.), Computer Vision - {ECCV} 2018 -
  15th European Conference, Munich, Germany, September 8-14, 2018, Proceedings,
  Part {XI}, Vol. 11215, pp. 793--810.

\bibitem{ALZSDA}
J.~Wang, J.~Jiang, \href{https://arxiv.org/abs/2009.05214}{Adversarial learning
  for zero-shot domain adaptation}, CoRR abs/2009.05214.
\newblock \href {http://arxiv.org/abs/2009.05214} {\path{arXiv:2009.05214}}.
\newline\urlprefix\url{https://arxiv.org/abs/2009.05214}

\bibitem{CausalityML}
B.~Sch{\"{o}}lkopf, Causality for machine learning, CoRR abs/1911.10500.

\bibitem{yu2020causality}
K.~Yu, X.~Guo, L.~Liu, J.~Li, H.~Wang, Z.~Ling, X.~Wu, Causality-based feature
  selection: Methods and evaluations, ACM Computing Surveys (CSUR) 53~(5)
  (2020) 1--36.

\bibitem{RCIT}
K.~Kuang, B.~Li, P.~Cui, Y.~Liu, J.~Tao, Y.~Zhuang, F.~Wu, Stable prediction
  via leveraging seed variable, CoRR abs/2006.05076.

\bibitem{BAMB}
Z.~Ling, K.~Yu, H.~Wang, L.~Liu, W.~Ding, X.~Wu, {BAMB:} {A} balanced markov
  blanket discovery approach to feature selection, {ACM} Trans. Intell. Syst.
  Technol. 10~(5) (2019) 52:1--52:25.

\bibitem{S2TMB}
T.~Gao, Q.~Ji, Efficient score-based markov blanket discovery, Int. J. Approx.
  Reason. 80 (2017) 277--293.

\bibitem{LLXDHT20}
S.~Li, C.~H. Liu, Q.~Lin, B.~Xie, Z.~Ding, G.~Huang, J.~Tang, Domain
  conditioned adaptation network, in: The Thirty-Fourth {AAAI} Conference on
  Artificial Intelligence, February 7-12, New York, NY, USA, 2020, pp.
  11386--11393.

\bibitem{SSRLDA}
S.~Yang, H.~Wang, Y.~Zhang, P.~Li, Y.~Zhu, X.~Hu, Semi-supervised
  representation learning via dual autoencoders for domain adaptation, Knowl.
  Based Syst. 190 (2020) 105161.

\bibitem{WeiFeature}
P.~Wei, Y.~Ke, C.~K. Goh, Feature analysis of marginalized stacked denoising
  autoenconder for unsupervised domain adaptation, IEEE transactions on neural
  networks and learning systems 30~(5) (2019) 1321--1334.

\bibitem{SERA}
S.~Yang, Y.~Zhang, H.~Wang, P.~Li, X.~Hu, Representation learning via serial
  robust autoencoder for domain adaptation, Expert Systems with Applications.
  (2020) 113635.

\bibitem{SEAE}
S.~Yang, Y.~Zhang, Y.~Zhu, P.~Li, X.~Hu, Representation learning via serial
  autoencoders for domain adaptation, Neurocomputing 351 (2019) 1--9.

\bibitem{WangJLWJ19}
X.~Wang, Y.~Jin, M.~Long, J.~Wang, M.~I. Jordan, Transferable normalization:
  Towards improving transferability of deep neural networks, in: Advances in
  Neural Information Processing Systems 32: Annual Conference on Neural
  Information Processing Systems, Vancouver, BC, Canada, December 8-14, 2019,
  pp. 1951--1961.

\bibitem{GhifaryKZB15}
M.~Ghifary, W.~B. Kleijn, M.~Zhang, D.~Balduzzi, Domain generalization for
  object recognition with multi-task autoencoders, in: 2015 {IEEE}
  International Conference on Computer Vision, {ICCV} 2015, Santiago, Chile,
  December 7-13, 2015, 2015, pp. 2551--2559.

\bibitem{LiPWK18}
H.~Li, S.~J. Pan, S.~Wang, A.~C. Kot, Domain generalization with adversarial
  feature learning, in: 2018 {IEEE} Conference on Computer Vision and Pattern
  Recognition, {CVPR} 2018, Salt Lake City, UT, USA, June 18-22, 2018, 2018,
  pp. 5400--5409.

\bibitem{DouCKG19}
Q.~Dou, D.~C. de~Castro, K.~Kamnitsas, B.~Glocker, Domain generalization via
  model-agnostic learning of semantic features, in: Advances in Neural
  Information Processing Systems 32: Annual Conference on Neural Information
  Processing Systems 2019, NeurIPS 2019, December 8-14, Vancouver, BC, Canada,
  2019, pp. 6447--6458.

\bibitem{MagliacaneOCBVM18}
S.~Magliacane, T.~van Ommen, T.~Claassen, S.~Bongers, P.~Versteeg, J.~M. Mooij,
  Domain adaptation by using causal inference to predict invariant conditional
  distributions, in: Advances in Neural Information Processing Systems 31,
  December 3-8, Canada, 2018, pp. 10869--10879.

\bibitem{RojasCarullaST18}
M.~Rojas{-}Carulla, B.~Sch{\"{o}}lkopf, R.~E. Turner, J.~Peters, Invariant
  models for causal transfer learning, J. Mach. Learn. Res. 19 (2018)
  36:1--36:34.

\bibitem{DGBR}
K.~Kuang, P.~Cui, S.~Athey, R.~Xiong, B.~Li, Stable prediction across unknown
  environments, in: Proceedings of the 24th {ACM} {SIGKDD} International
  Conference on Knowledge Discovery {\&} Data Mining, {KDD} 2018, August 19-23,
  London, UK, 2018, pp. 1617--1626.

\bibitem{Shen0ZK20}
Z.~Shen, P.~Cui, T.~Zhang, K.~Kuang, Stable learning via sample reweighting,
  in: The Thirty-Fourth {AAAI} Conference on Artificial Intelligence, {AAAI}
  2020, February 7-12, New York, USA,, 2020, pp. 5692--5699.

\bibitem{CRLR}
Z.~Shen, P.~Cui, K.~Kuang, B.~Li, P.~Chen, Causally regularized learning with
  agnostic data selection bias, in: 2018 {ACM} Multimedia Conference on
  Multimedia Conference, {MM} 2018, October 22-26, Seoul, Republic of Korea,
  2018, pp. 411--419.

\bibitem{KuangX0A020}
K.~Kuang, R.~Xiong, P.~Cui, S.~Athey, B.~Li, Stable prediction with model
  misspecification and agnostic distribution shift, in: The Thirty-Fourth
  {AAAI} Conference on Artificial Intelligence, February 7-12, New York, NY,
  USA, 2020, pp. 4485--4492.

\bibitem{YuL04}
L.~Yu, H.~Liu, Efficient feature selection via analysis of relevance and
  redundancy, J. Mach. Learn. Res. 5 (2004) 1205--1224.

\bibitem{PengLD05}
H.~Peng, F.~Long, C.~H.~Q. Ding, Feature selection based on mutual information:
  Criteria of max-dependency, max-relevance, and min-redundancy, {IEEE} Trans.
  Pattern Anal. Mach. Intell. 27~(8) (2005) 1226--1238.

\bibitem{TsamardinosA03}
I.~Tsamardinos, C.~F. Aliferis, Towards principled feature selection:
  Relevancy, filters and wrappers, in: C.~M. Bishop, B.~J. Frey (Eds.),
  Proceedings of the Ninth International Workshop on Artificial Intelligence
  and Statistics, {AISTATS} 2003, Key West, Florida, USA, January 3-6, 2003.

\bibitem{AliferisTS03}
C.~F. Aliferis, I.~Tsamardinos, A.~R. Statnikov, {HITON:} {A} novel markov
  blanket algorithm for optimal variable selection, in: {AMIA} 2003, American
  Medical Informatics Association Annual Symposium, Washington, DC, USA,
  November 8-12, 2003, 2003.

\bibitem{NiinimakiP12}
T.~Niinimaki, P.~Parviainen, Local structure discovery in bayesian networks,
  in: Proceedings of the Twenty-Eighth Conference on Uncertainty in Artificial
  Intelligence, Catalina Island, CA, USA, 14-18 August, 2012, pp. 634--643.

\bibitem{neufeld1993pearl}
J.~Pearl, Probabilistic reasoning in intelligent systems: networks of plausible
  inference, Morgan Kaufmann series in representation and reasoning, 1988.

\bibitem{CPS}
P.~Spirtes, C.~Glymour, R.~Scheines, Causation, Prediction, and Search, {MIT}
  Press, 2000.

\bibitem{ZhengARX18}
X.~Zheng, B.~Aragam, P.~Ravikumar, E.~P. Xing, Dags with {NO} {TEARS:}
  continuous optimization for structure learning, in: Advances in Neural
  Information Processing Systems 31: Annual Conference on Neural Information
  Processing Systems, December 3-8, Montr{\'{e}}al, Canada, 2018, pp.
  9492--9503.

\bibitem{SunFS16}
B.~Sun, J.~Feng, K.~Saenko, Return of frustratingly easy domain adaptation, in:
  D.~Schuurmans, M.~P. Wellman (Eds.), Proceedings of the Thirtieth {AAAI}
  Conference on Artificial Intelligence, February 12-17, Phoenix, Arizona, USA,
  2016, pp. 2058--2065.

\bibitem{WangCYHY19}
J.~Wang, Y.~Chen, H.~Yu, M.~Huang, Q.~Yang, Easy transfer learning by
  exploiting intra-domain structures, in: {IEEE} International Conference on
  Multimedia and Expo, July 8-12, Shanghai, China, 2019, pp. 1210--1215.

\bibitem{CaoLW18}
Y.~Cao, M.~Long, J.~Wang, Unsupervised domain adaptation with distribution
  matching machines, in: Proceedings of the Thirty-Second {AAAI} Conference on
  Artificial Intelligence, February 2-7, New Orleans, USA, 2018, pp.
  2795--2802.

\bibitem{Demsar06}
J.~Demsar, Statistical comparisons of classifiers over multiple data sets, J.
  Mach. Learn. Res. 7 (2006) 1--30.

\end{thebibliography}
\end{sloppypar}
\end{document}